\documentclass{article}
\usepackage{microtype}
\usepackage{graphicx}
\usepackage{subfigure}
\usepackage{booktabs} 
\usepackage{hyperref}
\usepackage[accepted]{icml2023}
\usepackage{amsmath}
\usepackage{amssymb}
\usepackage{mathtools}
\usepackage{amsthm}
\usepackage[capitalize,noabbrev]{cleveref}
\usepackage[utf8]{inputenc}
\usepackage[T1]{fontenc}
\usepackage{url}
\usepackage{booktabs}
\usepackage{algorithmic}
\usepackage{algorithm}
\usepackage{amsfonts}
\usepackage{nicefrac}
\usepackage{xcolor}
\usepackage{graphicx}
\usepackage{wrapfig}

\DeclareMathOperator*{\argmin}{arg\,min}

\DeclareMathOperator{\rgt}{\textbf{rgt}}

\theoremstyle{plain}
\newtheorem{theorem}{Theorem}[section]

\newtheorem{lemma}[theorem]{Lemma}

\theoremstyle{definition}

\theoremstyle{remark}

\icmltitlerunning{Neural Auctions Compromise Bidder Information\hfill\thepage}

\begin{document}

\twocolumn[
\icmltitle{Neural Auctions Compromise Bidder Information}
\icmlsetsymbol{equal}{*}

\begin{icmlauthorlist}
\icmlauthor{Alex Stein}{equal,umd}
\icmlauthor{Avi Schwarzschild}{equal,umd}
\icmlauthor{Michael Curry}{zurich}
\icmlauthor{Tom Goldstein}{umd}
\icmlauthor{John Dickerson}{umd}
\end{icmlauthorlist}

\icmlaffiliation{umd}{University of Maryland, College Park, USA}
\icmlaffiliation{zurich}{University of Zurich, Zurich, Switzerland}
\icmlcorrespondingauthor{Alex Stein}{astein0@umd.edu}
\icmlkeywords{Machine Learning, ICML}
\vskip 0.3in
]



\printAffiliationsAndNotice{\icmlEqualContribution} 

\begin{abstract}
  \looseness=-1
  Single-shot auctions are commonly used as a means to sell goods, for example when selling ad space or allocating radio frequencies, however devising mechanisms for auctions with multiple bidders and multiple items can be complicated.
  It has been shown that neural networks can be used to approximate optimal mechanisms while satisfying the constraints that an auction be strategyproof and individually rational.
  We show that despite such auctions maximizing revenue, they do so at the cost of revealing private bidder information.
  While randomness is often used to build in privacy, in this context it comes with complications if done without care. Specifically, it can violate rationality and feasibility constraints, fundamentally change the incentive structure of the mechanism, and/or harm top-level metrics such as revenue and social welfare.
  We propose a method that employs stochasticity to improve privacy while meeting the requirements for auction mechanisms with only a modest sacrifice in revenue.  We analyze the cost to the auction house that comes with introducing varying degrees of privacy in common auction settings.  Our results show that despite current neural auctions' ability to approximate optimal mechanisms, the resulting vulnerability that comes with relying on neural networks must be accounted for. 
\end{abstract}

\section{Introduction}
\label{sec:intro}

An auction is an economic mechanism that elicits bids from bidders and allocates goods in exchange for a fee. In the private-value auction setting, bidder valuations are drawn from some known prior probability distribution, while their true valuations are kept private from the mechanism and other bidders.  Therefore, when the auctioneer elicits bids, the bidders may bid strategically by misreporting their valuations in order to benefit themselves. The auctioneer's goal is to design an auction that has desirable properties in the face of such strategic behavior.

Auctions designers generally want mechanisms to ensure two properties: auctions should be strategyproof and maximize revenue in expectation. An auction mechanism is considered strategyproof when bidders are no better off bidding anything other than their true valuations for the goods.  Given this hard constraint of strategyproofness, the auctioneer would want to maximize expected revenue.  However, strategyproof auctions inherently leak information about a bidders true valuation; by incentivizing bidders to bid truthfully, it might be possible to learn private bidder information by observing the resulting allocations and payments of a strategyproof auction.

Finding optimal auction mechanisms has been a subject of great interest to economists for decades.  The Myerson auction \citep{Myerson1981Optimal} is optimal for selling one item to multiple bidders. Additionally, for selling multiple items to one bidder, optimality is also somewhat well-understood \citep{Manelli2006Bundling,Pavlov2011Optimal,Daskalakis2017Strong,Kash2016Optimal}. However, for selling multiple items to multiple bidders, optimal mechanisms are known only in very simple cases~(e.g., \citet{Yao2017Dominant}). Even for selling two items to two bidders with i.i.d. uniform valuations, the optimal mechanism is unknown. 

The apparent intractability of finding analytic results is what motivates the work of~\citet{dutting2019optimal}. Their neural network approximations of optimal auction mechanisms are not provably revenue maximizing or strategyproof, however they approximately recover optimal mechanisms of known auctions and, in settings without known optimal mechanisms, their revenues are higher than existing alternatives and the associated regret (a measurement of strategyproofness) is sufficiently small.

The strategyproofness and revenue maximizing properties of these ``neural auctions'' are reasonably well understood, but bidders may also be concerned with privacy. If the mechanism is strategyproof, bidders are incentivized to reveal their true valuations to the mechanism itself, but they may still want their private valuations not to be revealed to other bidders or outside observers. For example, a business may not want its willingness to pay for different items to be known to competitors or counterparties. Yet if the mechanism's rules, allocations, and payments are made public, it might be possible to infer a large amount of information about a bidder's bid.

In this work we analyze neural auctions of several bidder-item sizes and show that they are not private. In particular, using the outputs of the networks -- the publicly available allocation and payment information -- a malicious actor can invert the models to extract information about agents' valuations. Optimization-based model inversion is a common security vulnerability of neural models used for tasks ranging from facial recognition \citep{fredrikson2015model} to natural language generation \citep{pan2020privacy}. In most settings, however, the fear of the practitioner is that the training data is retrievable. Our case deviates slightly from current analysis in that we are concerned with protecting single input examples at test time.

After demonstrating that the mechanisms described in prior work are not private in this sense, we propose a technique to introduce privacy. We employ stochasticity by adding noise to the outputs to create private neural auctions, which we show empirically mitigates model inversion. This randomization process is non-trivial, as the resulting network needs to satisfy rationality and strategyproofness constraints. Furthermore, this method introduces a parameter -- the standard deviation of the noise -- to control the degree of privacy. It is critical to give auction designers this control since privacy comes at a cost to revenue for the auction house. We analyze this cost by characterizing the relationship between privacy and revenue and regret.

\section{Related Work}
\label{sec:related-work}

\paragraph{Differentiable economics}

Differentiable economics is an approach to automated mechanism design that uses rich function approximators to learn good mechanisms \citep{Sandholm2003Automated}.  \citet{dutting2019optimal} propose RegretNet, a neural auction, and show that neural networks can approximate optimal auction mechanisms. While our work builds directly on theirs, there have been a number of other recent contributions to topics within differentiable economics (many of which also build on the work). 

\citet{Rahme2021Auction} expand on RegretNet, by proposing a new training algorithm for similar auction networks as well as providing a metric for measuring performance. \citet{feng2018deep} extend the RegretNet framework to consider private budget constraints as well as proving optimality in some specific and solved auction settings. \citet{duan2022context} and \citet{ivanov2022optimal} use attention layers to produce networks which are permutation equivariant, generalize to unseen data, and perform better. These extensions focus on revenue maximization for nearly regret-free auctions, but do not consider provably strategyproof settings. Furthermore, \citet{Curry2020Certifying} introduce explicit certificates for the degree to which strategyproofness is violated.

Some architectures are successful at finding perfectly strategyproof auctions, however each implementation only applies in limited settings.  In the same paper that introduces RegretNet, \citet{dutting2019optimal} propose RochetNet, which is strategyproof by construction but is restricted to the single bidder setting.  Similarly, \citet{Shen2019Automated} design MenuNet, a framework which also learns perfectly strategyproof auctions, but is limited to the single agent setting. Additionally, in settings with various bidder demands, mechanisms can be learned with differentiable matching \citep{Curry2021Learning}. While these models consider only the single bidder setting, \citet{CurryAMA2022} designs Affine Maximizer Auctions (AMAs) that are provably strategyproof and handle multi-agent auctions, however they are not necessarily revenue maximizing.

Other forms of mechanism design beyond auctions have been explored as well.  Early work by \citet{narasimhan2016automated} sets the stage for RegretNet by exploring the use of machine learning to learn strategyproof mechanisms that optimize social choice problems. Additionally, they introduce the concept of optimizing stability within two-sided matching problems.  \citet{ravindranath2021deep} further study the matching problem and quantify the trade-off between strategyproofness and stability. Finally, \citet{golowich2018deep} use deep learning to find strategyproof models for minimizing expected social cost within multi-facility location mechanisms.

\paragraph{Privacy in mechanism design}
There is a longstanding thread of research on privacy in mechanism design theory -- see~\citet{pai2013privacy} for a survey and introduction. The typical aim  of privcay tools for auctions is to guarantee a differential privacy (DP) property with respect to the bids. Of particular relevance is work that constructs mechanisms satisfying this property by averaging over random noise, which shows that the resulting smoothness gives both privacy and some notion of strategyproofness~\citep{mcsherry2007mechanism}.

We do not directly deal with the differential privacy formalism -- instead we view privacy from the perspective of model inversion attacks in deep learning. Importantly, DP is concerned with recovering training data and DP training algorithms aim to find models that would be no different with or without the inclusion of small amounts of data \citep{Abadi_2016}. However, in practice, auctions can (and are) trained with synthetic data and the exact training data can be made public without breaching any personal privacy. The issue at hand in this paper is about inferring inputs at test time from publicly available outputs – not a common issue in other domains. We do, however, make use of random noise, which has some connections to DP~\citep{lecuyer2019certified}. 

The setting we analyze includes outputs that are publicly available after the completion of the auction. This also stands in contrast to cryptographically secure markets such as dark pools and transactions that rely on secure multi-party computation (MPC). \citet{stockmarketmpc} uses MPC to implement a framework for stock market participants to avoid a potentially untrustworthy auctioneer.  For analysis on using MPC to privately clear transactions in continuous double auctions see ~(e.g., \citet{mpcjoinsdarkside, privacypreservingpool}).

\paragraph{Model Inversion}
Inverting machine learning models to recover data used for training has been studied extensively. ~\citet{model_inversion} formalize a framework for analyzing model inversion in both the black box (oracle access) and white box (where the adversary has access to the composition of the model) settings. ~\citet{adversarial_attack} show that an adversary can (with black box access to the model)  train an inversion model to effectively target a machine learning model by augmenting the inversion training with auxiliary data.

While these works discuss the inversion of neural networks, the auction setting is unique in that the test-time outputs are public and the adversary is attempting to recover test-time inputs, rather than inverting the model to learn train-time data.  In the mechanism design literature, we generally assume white box access to the model for all participants.  Additionally, we are unaware of any prior work focusing on understanding the trade-off between model privacy and cost in terms of revenue.

\section{Problem Setting}
\label{sec:background}

For ease of discussion, we define the problem of finding an optimal auction and review common notation. Consider an auction consisting of a set of $n$ bidders $N=\{1, ..., n\}$ and a set of $m$ items $M=\{1, ..., m\}$ where each bidder $i$ has some private valuation $v_i \geq 0$ drawn from some distribution $P_i$ over $V_i$ the possible valuations for each of the $m$ goods.  In general, bidders can either have \emph{unit demand}, where their valuation of $S$ items is $v_i(S) = \max_{j \in S} v_i(j)$, or they can have \emph{additive demands}, where each bidder wants as many of the items as possible. The bidders each present bids to the auction mechanism and these bids may or may not truthfully represent their valuations.

After eliciting bids from each of the $n$ agents, the auction allocates the $m$ goods to the bidders charging each of them some payment. Formally, define an auction as $f = (g,p)$ or a set of allocation and payment rules $g(b)$ and $p(b)$ that take as input the bidders' bids $b = (b_1, b_2, ... b_n)$. An auction acts in its best interest, which is often some form of revenue maximization (more detail below).

Each bidder is trying to maximize their own \textit{utility}, $u_i(v_i;b) = v_i(g_i(b)) - p_i(b_i)$, where $v_i$ is their valuation, $b$ is the bid profile of all the bids, and $g_i, p_i$ are the allocations and payments for bidder $i$ given the bid profile $b$. Also, let $b_{-i}$ be the bid profile without bidder $i$. If bidder utility is maximized by bidding truthfully \emph{even without knowledge of how other bidders are acting}, then this is considered a \emph{dominant strategy incentive compatible} (DSIC), or strategyproof, auction. Formally, DSIC holds if the following is satisfied.
\begin{equation}
    \forall i\, \forall v, b:\, u_i(v_i;v_i, b_{-i}) \geq u_i(v_i;b_i, b_{-i}).
\end{equation}

Identifying a DSIC set of rules $(g, p)$ that maximize revenue in expectation is an open problem for multi-agent multi-item auctions. To address this, \citet{dutting2019optimal} propose RegretNet -- a neural network that can learn a near optimal mechanism from data. Recent advances in deep learning for auctions include ALGNet \citep{Rahme2021Auction}, which uses a modified training algorithm, as well as RegretFormer \citep{ivanov2022optimal} and CITransNet \citep{duan2022context}, both of which use self-attention layers to make mechanisms that are expressive and permutation-equivariant. What all of these approaches have in common is that they parameterize the auction with a neural network. In this paper, we focus on the original RegretNet approach, where the network backbone is a simple multi-layer perceptron.  
 
\subsection{Learning problem} 
To explicitly formulate strategyproof auctions as a deep learning problem in terms of minimizing regret, we must first define the concepts of regret and bidder utility.  An auction $f = (g,p)$ can be parameterized $f(b; \theta) = [g(b), p(b)]$, where $\theta$ denotes the trainable parameters. Let bidder $i$ have a utility function defined as follows -- the \emph{welfare} from items received, minus the \emph{payment}.
\begin{equation}
    \label{def:bidder_utility}
    u_i^\theta(v_i;b) = v_i(g_i^\theta(b)) - p_i^\theta(b)
\end{equation}
Using this definition of bidder utility, we can define a bidder's regret as the difference in utility between bidding truthfully and bidding to maximize utility. For bidder $i$, $\rgt$ as a function of the private valuations $v$ is defined as follows.
\begin{equation}
    \label{def:regret}
    \rgt_i(v) = \max_{b_i \in V_i} u_i^\theta(v_i;(b_i, v_{-i})) - u_i^\theta(v_i;(v_i, v_{-i}))
\end{equation}
Where $\rgt$ without a specific index is the expected value of $\rgt_i$ for all $i$, which we compute as the average regret value over a large sample of inputs. Also, agents are subject to individually rationality (IR), and would not participate in an auction that could lead to negative utility.

Finally, the optimization problem to find a feasible, strategyproof, revenue-maximizing auction can be formulated by minimizing the expected negated total payment subject to no-regret and IR.
\begin{equation}
    \begin{aligned}
        \min_{\theta\in \mathbb{R}^d} & \quad  \underset{v \sim P}{\mathbb{E}} \left[-\sum_{i\in N} p_i^\theta(v)\right] \\
        \text{s.t.} \rgt_i(v) & \approx 0, \forall i\in N, v\\
        u_i(v) & \geq 0\\
        \sum_{i\in N} g_{i,j}(v) & \leq 1, \forall j \in M
    \end{aligned}
\end{equation}
The neural architecture we use to solve this problem has three components.
The \emph{backbone} is a multi-layer perceptron with ReLU activations where the depth and width are specified per auction size below. This component takes as input the $n\times m$ bid array (bids from all bidders for all items) and produces a feature vector. The other two components of the network are output modules, one for allocations and one for payments. The \emph{allocation module} takes the feature vector and passes it through a single fully connected layer to produce an $(n+1) \times m$ output. We compute an item-wise softmax over the allocations to ensure that no more than one unit of each item is allocated. Note, the extra row in the allocation output allows for allocating less than a whole item to the bidders, accounting for times when the auction house may not sell the entire good. The \emph{payment module} is similar, consisting of one fully connected layer to map the feature vector to a vector of length $n$, to which we apply a entry-wise sigmoid. These values correspond to payments, which are a fraction of bidder welfare. The actual fee each agent is charged is computed by multiplying their payment fraction by their welfare.

To estimate the degree of regret, there is an inner loop during training, performing gradient steps on the inputs to approximately maximize the utility from misreporting. Given these basic design choices, RegretNet maximizes payment and includes the (estimated) regret in an augmented Lagrangian term to enforce the constraints. For the complete training details, see the original paper \citep{dutting2019optimal}. After training, we have allocation and payment networks that have very low regret (so the mechanism is very nearly DSIC).

\subsection{Performance Baselines}
\label{sec:performance_baselines}

To contextuallize the benefit of using neural networks, it is critical to study the alternatives. Without learning auctions from data, one might execute independent Myerson auctions. For $m$ items, it is possible to run $m$ independent Myerson auctions, preserving strategyproofness. While these are not revenue-optimal, they at least have revenue which is optimized separately for each item, and in the limit of the number of bidders they approach optimality \citep{Palfrey1983Bundling} so they are useful baseline for learned mechanisms.

An alternative baseline for strategyproof auctions is AMAs. ~\citet{CurryAMA2022} employ Lottery AMAs to improve revenue compared to the item-wise Myerson auction, while remaining fully strategyproof.  These auctions are not revenue-optimal but as opposed to other possible baselines (i.e., MenuNet, RochetNet, etc) AMAs are applicable to the multi-agent setting.  

\begin{table}[ht!]
    \centering
    \caption{\textbf{Average revenue and regret for the mechanisms we consider.} All agents have additive valuations drawn from $U[0,1]$. Myerson Auctions have 0 regret by construction.  For RegretNet metrics, we show averages over ten random seeds.}
    \label{tab:revenue-regret}
    \small
    \begin{tabular}{cccccc}
    \toprule
     & \multicolumn{2}{c}{RegretNet} && \multicolumn{2}{c}{Myerson} \\ 
    \cmidrule{2-3} \cmidrule{5-6}
     Size                &   Revenue    & Regret && Revenue \\
                     \midrule
     $1 \times 2$    & $0.5720$ &   $0.0011$ && $0.500$\\
     $2 \times 2$    & $0.8818$ &   $0.0008$ && $0.833$\\
     $3 \times 2$    & $1.1284$ &   $0.0028$ && $1.063$\\
     $2 \times 3$    & $1.3522$ &   $0.0040$ && $1.250$\\
     $3 \times 3$    & $1.6837$ &   $0.0046$ && $1.594$\\
     \bottomrule
    \end{tabular}
\end{table}

Table \ref{tab:revenue-regret} shows the revenue and regret of the mechanisms we study. These figures are consistent with the notion that without considering privacy, RegretNet provides the auction house with more revenue than our baseline auctions and is nearly DSIC. We study the most common setting, where bidders draw valuations independently from $U[0,1]$. See Appendix \ref{sec:app-training} for additional training hyperparameters and model architecture details.
For further detail we show the revenue and regret numbers for different baselines for the two agent, two item setting in Table ~\ref{tab:baselines}.

\begin{table}[ht!]
    \centering
    \caption{\textbf{Revenue for $2 \times 2$ auctions for different baselines as compared to RegretNet.} Baselines with zero regret are zero regret by construction}
    \label{tab:baselines}
    \small
    \begin{tabular}{lcc}
        \toprule
        Auction & Revenue & Regret\\
                         \midrule
        Item-Wise Myerson                   & $0.833$ & 0\\
        Lottery AMA~\cite{CurryAMA2022}     & $0.868$ & 0 \\
        ALGNet~\cite{Rahme2021Auction}      & $0.879$ & $<0.001$ \\
        RegretNet~\cite{dutting2019optimal} & $0.882$ & $<0.001$ \\
        \bottomrule
    \end{tabular}
\end{table}

\section{Privacy in Auctions}
\label{sec:not-private}

The central line of inquiry in our work aims in part to answer the question: Given the auction mechanism and knowledge of the results of a particular auction, how easy is it to recover bidders' private valuations using model inversion techniques?

We show that in a neural auction, an outside observer can, in fact, retrieve information about the bids. Furthermore, we show that a participant in the auction, who may use their own bid information in the process, can recover even more information about other bidders. This vulnerability renders these mechanisms flawed in settings where bidder information must remain private. 

\subsection{Threat Model}

We consider two cases where the adversary has varying degrees of information at their disposal. 
First, we treat the case where an outside observer with no knowledge of any bids attempts to invert the model after observing the allocations and payments.
We also look at the situation where one of the bidders, who knows their own bid, aims to recover other bidders' information.
In both cases, we assume that the mechanism itself (i.e. the network weights) is publicly available -- this is a reasonable assumption as it is typically assumed that auction participants know what the mechanism is.

An adversary can attempt to guess bids $x^*$ by selecting the inputs that produce payments and allocations close to the true payments and allocations. It is important to note that this strategy assumes that for a given set of payment and allocation outputs, there is only one input -- or that the model is a bijective function. This assumption is strong for general mechanisms, but in practice we find \textit{neural} auctions to be invertible suggesting that it is often the case.

Therefore, identifying $x^*$ by solving the optimization problem below is akin to finding the true bids $b$.
\begin{equation}
    x^* = \argmin_{x \in \text{supp}(P)} \|g(x) - g(b)\|_{\ell_2} + \|p(x) - p(b)\|_{\ell_2}
    \label{eq:model-inv}
\end{equation}
As is standard in the auction literature, our implementation assumes the adversary knows the distribution $P$ from which agents draw their valuations -- this is utilized by initializing the guess to be a random sample from that distribution (see \ref{sec:app-inversion} for more details).  However it is worth mentioning that even in settings without a view about the bid generating distribution, an adversary that initializes their bids with zeros is able to successfully learn private information about bids (see Appendix~\ref{sec:app-no-access}) Note that in the setting where the adversary is also participating in the mechanism as a bidder, the free variables in the Equation~\eqref{eq:model-inv} correspond only to the other bidders' bids.

\subsection{Privacy Metrics and Baselines}
\label{sec:metrics}

In order to quantify how successful an adversary is able to invert an auction we introduce privacy metrics and baselines. We measure the adversary's success in two ways. The first is by computing the percentage of recovered bids that are within some tolerance of the true bids. The second method is to compute the average absolute error in the bid reconstruction over some large set of examples.  Recall that our focus is the setting in which the adversary knows the distribution from which agents draw their bids.
%
Complete privacy would ensure that our inversion is more accurate than a random draw from the bid generating distribution.  However, a more applicable goal would be to beat the privacy of the item-wise Myerson auctions, since they often serve as a baseline for comparison for learned mechanisms. Because we know that Myerson auctions are exactly strategyproof, any optimal neural auction must be at least as private as a Myerson auction to be strictly more desirable.  See Appendix~\ref{sec:myerson} for a discussion of the privacy of Myerson auctions.

\subsection{Neural Auctions Are Not Private}
\label{sec:results}

We examine auctions of five sizes and for each, we show how much information the adversary can extract. When the mechanism is learned, and therefore differentiable, the adversary uses gradient-based optimization to invert the model. Specifically, we execute gradient descent to solve the problem in Equation \eqref{eq:model-inv}, projecting the bids onto $[0, 1]^{mn}$ at each iteration.  Section~\ref{sec:app-inversion} gives the model inversion algorithm.  It is important to note that as opposed to most cases of model inversion that focus on the setting where an adversary is trying to learn training inputs, here the adversary cares about learning test inputs.  This separates our analysis from prior work on the vulnerabilities of neural networks.

Table~\ref{tab:recovery-rates} shows the recovery rates for several sizes of auctions at both levels of information access. We measure the recovery rates here as a percentage of bids that were successfully extracted from the model inversion to within a tolerance of $\pm 0.02$.  This cutoff was chosen arbitrarily but serves mainly as a way to demonstrate being ``close'' to exact.  For a distance measure, we also included the absolute error of the prediction in table~\ref{tab:recovery-error}. These figures do not give the adversary any guarantees. In particular, we do not claim a bound on the error for any given entry in their reconstructed bid matrix. However, this is still a major vulnerability for the auction house since the adversary has a high likelihood that their estimate of any particular bid is correct.  While the figures below are for the case where the adversary knows that the bids are drawn from a uniform distribution $P_i\sim U[0,1]$ in Section~\ref{sec:app-no-access} we show similar results even when the adversary has no information about $P$ (except that no types are negative) and instead initializes guesses to zeros.

Furthermore, we compare to the recovery rate one might get from randomly guessing bids from the known valuation distribution. With the figures in the last row of Table \ref{tab:recovery-rates}, we highlight just how much more information the adversary has as a direct result of inverting the mechanism.

\begin{table*}[t!]
    \centering
    \caption{\textbf{Bid recovery rates.} For different auction sizes, we show the portion of the bids (in percentage $\pm$ standard error) that the adversary can recover within a tolerance of $\pm0.02$. Note, for bids drawn i.i.d. from $U[0,1]$ a random guess sampled from the same distribution is within the tolerance $3.96\%$ of the time. Whether the adversary has any true bidder information or not, their recovery rate is well above random.}
    \label{tab:recovery-rates}
    \footnotesize
    \begin{tabular}{lcccccccccc}
    \toprule
                        & \multicolumn{5}{c}{Bidders $\times$ Items} \\
                        & 1 $\times$ 2     & 2 $\times$ 2   & 3 $\times$ 2     & 2 $\times$ 3      & 3 $\times$ 3     \\
    \midrule
    W/ bids             & --               & $84.98\pm0.50$ & $82.62 \pm 0.39$ & $91.44 \pm 0.45$  & $82.97 \pm 6.58$ \\
    W/O any bids        & $77.92 \pm 1.47$ & $93.08\pm0.64$ & $78.27 \pm 0.60$ & $25.69 \pm 1.40$  & $53.53 \pm 4.15$ \\
    \bottomrule
    \end{tabular}
\end{table*}

Recovery rates are a relatable figure and for this reason we choose to describe the adversary's success this way, but the results in Table~\ref{tab:recovery-rates} are dependant on the tolerance we set. For metrics that are perhaps less relatable but that do not depend on any hand-picked tolerance, we look at the mean absolute error (MAE). Specifically, Table~\ref{tab:recovery-error} shows the average distance from the adversary's reconstruction to the true bid value. Whether considering MAE or recovery rates, it is clear that inverting neural auctions to uncover private bidder information is a real privacy vulnerability.

\begin{table}[ht!]
    \centering
    \vspace{-8pt}
    \caption{\textbf{Bid reconstruction error.} Adversary's MAE for different auctions. Note, for random guess sampled from the same distribution as the valuations, the MAE is $\frac13$ on average. Also, the standard error for every entry in this table is $<0.01$.}
    \label{tab:recovery-error}
    \small
    \begin{tabular}{lcccccccccccccc}
    \toprule
                        & \multicolumn{5}{c}{Bidders $\times$ Items} \\
                        & 1 $\times$ 2     & 2 $\times$ 2   & 3 $\times$ 2     & 2 $\times$ 3      & 3 $\times$ 3\\
    \midrule
    W/ bids             & --               & $0.02$         & $0.03$           &  $0.01$           & $0.021$      \\
    W/O any bids        & $0.03$           & $0.01$         & $0.03$           &  $0.10$           & $0.064$       \\
    \bottomrule
    \end{tabular}
\end{table}

\section{Non-Determinism for Privacy}
\label{sec:non-determinism}

With a clear picture of how vulnerable neural models are, we propose a method to prevent inversion. Our technique, which employs stochasticity to help obscure the inputs, comes at a cost to the auction house in expected revenue. We describe our defense to inversion attacks and we characterize the relationship between degrees of privacy and the cost to the auction house.

\begin{figure}[ht!]
    \centering
    \includegraphics[width=\columnwidth]{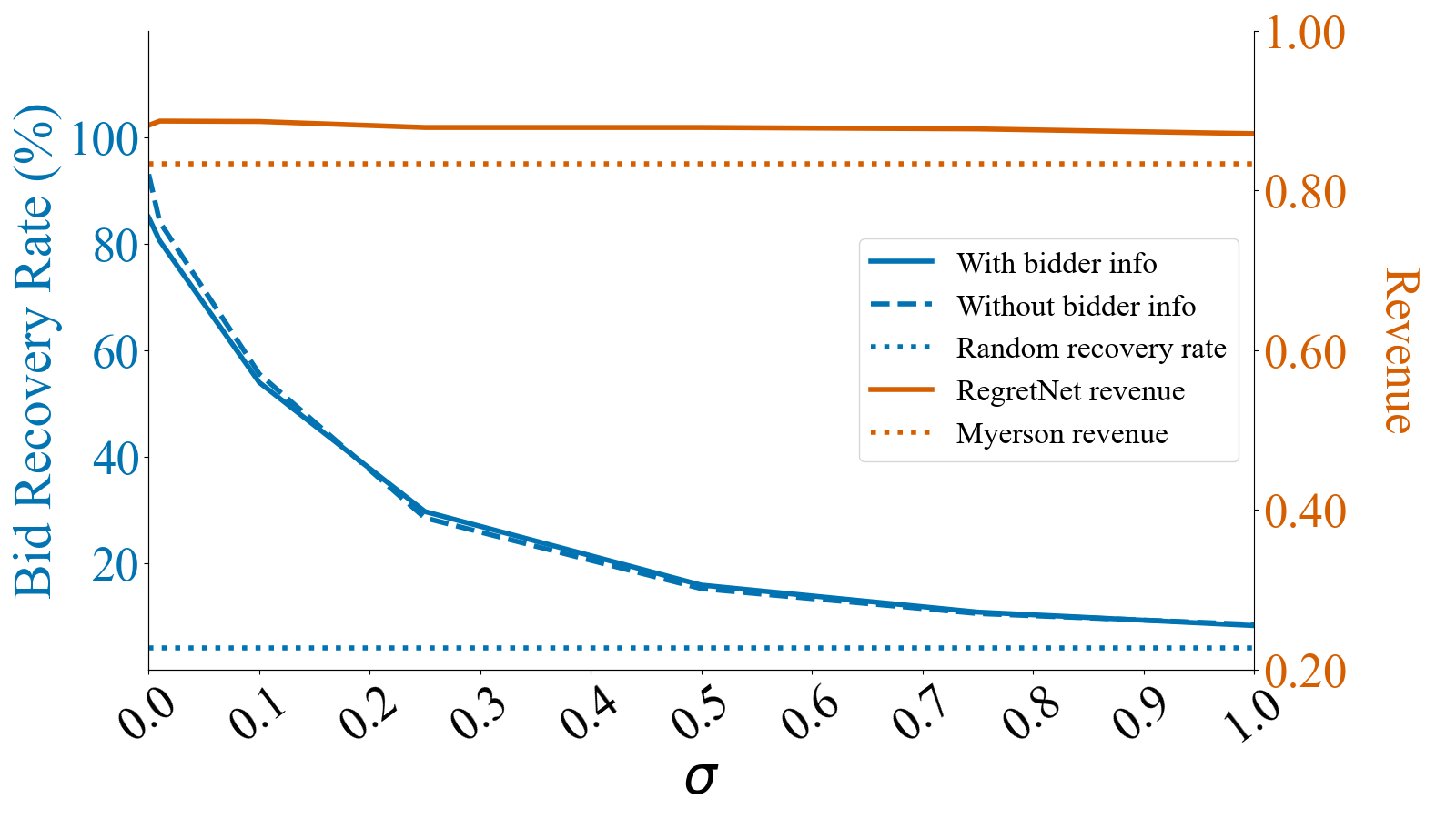}
    \caption{\textbf{Privacy and revenue in 2 $\times$ 2 auctions.} The recovery rate measures how often the inverted bids are within $\pm 0.02$ of the true values. With small $\sigma$ values that barely affect the expected revenue, we can mitigate model inversion. For example, note that when $\sigma = 0.2$, the revenue is steadily above the Myerson baseline but the recovery rate is cut in half.}
    \label{fig:cost-2x2}
\end{figure}

\subsection{Stochasticity That Satisfies Auction Constraints}
\label{sec:stochasticity}

Many previous works \citep[e.g.][]{Sandholm2003Automated} discuss stochasticity in mechanisms, often for the sake of computational efficiency. We use stochasticity to make inversion more difficult -- we perturb the allocations and payments output by the network. To satisfy individual rationality and ensure that at most one unit of each item is allocated, this perturbation requires care as follows.

RegretNet, by default, employs several architectural choices to meet the ex post individual rationality criteria. Specifically, the model first allocates all items to $n+1$ bidders, where the extra ``ghost'' bidder accounts for portions not sold. Using this ghost bidder allows the network to enforce, via softmax, that the allocations per item sum to one. We refer to the input to this softmax as \emph{allocation logits} and the output as \emph{allocations}. RegretNet also generates a payment factor, or a positive fraction less than one, and multiplies this by the value of items allocated to each bidder to compute payments that are guaranteed to satisfy ex post individual rationality. 

With this in mind, we choose to perturb the allocations matrix before the softmax is applied. If we perturb the allocations after the softmax, we risk entries in the allocation matrix being negative or the total allocation of a single item exceeding one unit.

We add independent mean zero Gaussian noise to each of the allocation logits and then use softmax to constrain the allocations matrix to have entries between zero and one and column-wise sums equal to one. The magnitude of this noise is controlled by the standard deviation $\sigma$ and we explore how this parameter affects the privacy metrics.

\subsection{Metrics of Randomized Auctions}
\label{sec:metrics_randomized}

Measuring the payment and regret of a model, for example when evaluating performance after training, requires estimating two expectations. First, we estimate the average regret over bidders (and samples).
\begin{equation}
    \rgt(\theta) = \underset{v\sim P}{\mathbb{E}} \left[\rgt(v)\right]
\end{equation}
 Second, we need to estimate the average payment, 
\begin{equation}
    \mathbf{p}(\theta) = \underset{v \sim P}{\mathbb{E}} \left[\sum_{i\in N} p_i^\theta(v)\right].
\end{equation}
Our models have stochasticity added ``inside'' the mechanism before the last layer. However, due to the use of linear, additive utilities, the expected welfare from winning items averaged over the stochasticity is the same as the welfare of the expected allocation.  That is to say that because the welfare is a linear function of the allocation, we can use the average allocation to compute the expected welfare without having to recompute the average welfare directly.  Thus, when estimating payments, we can simply average over a large number of sampled bid profiles with sampled noise, as usual.

To estimate regret under a single bid profile, we need to compute two utility terms in the regret for a given set of bids. We sample bid profiles and compute optimized misreports for each one. Then, given bids and misreports, we can compute expected allocations/payments in order to compute the two utility terms in Equation \eqref{def:regret}.

\subsection{Cost of Privacy}
\label{sec:cost_of_privacy}

For an auction house interested in carrying out private auctions, our stochastic technique comes with a trade-off. As the magnitude of the noise increases the auction becomes more private but it also realizes less revenue on average. Consider the $2\times2$ auction, where RegretNet offers bidders access to nearly all of the other bidders' bids. In Figure \ref{fig:cost-2x2}, we show that by adding a stochastic layer to the network whose noise has standard deviation $\sigma = 0.2$ the auction becomes twice as private (see the decay in the blue curves). Furthermore, with $\sigma = 1.0$, the adversary's bid recovery rate drops to around 10\%. By examining the orange curve in Figure \ref{fig:cost-2x2}, we see that the cost to the auction house is about 0.005 units or just over 0.5\%.

\begin{figure}[ht!]
    \centering
    \includegraphics[width=\columnwidth]{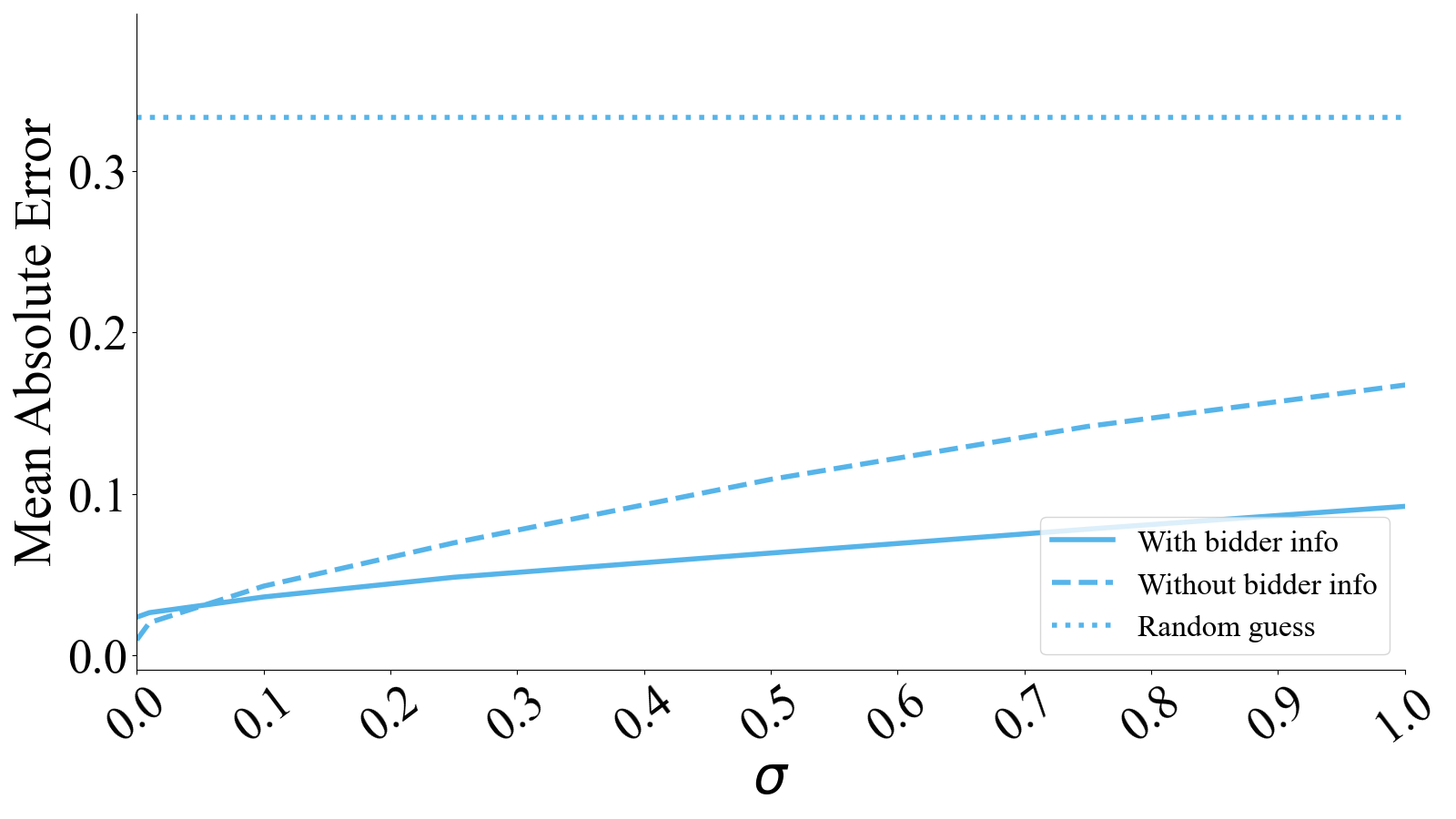}
    \caption{\textbf{Adversary's MAE in 2 $\times$ 2 auctions.} For a metric without a hand-picked parameter, we show MAE. Lower values correspond to more privacy. With this metric, it is clear that a lot of privacy can be gained with very little noise, but ensuring that the inversion is only as good as a random guess requires a large $\sigma$ value.}
    \label{fig:2x2-mae}
\end{figure}

\begin{figure}[ht!]
\begin{minipage}{\columnwidth}
    \centering
    \includegraphics[width=\textwidth]{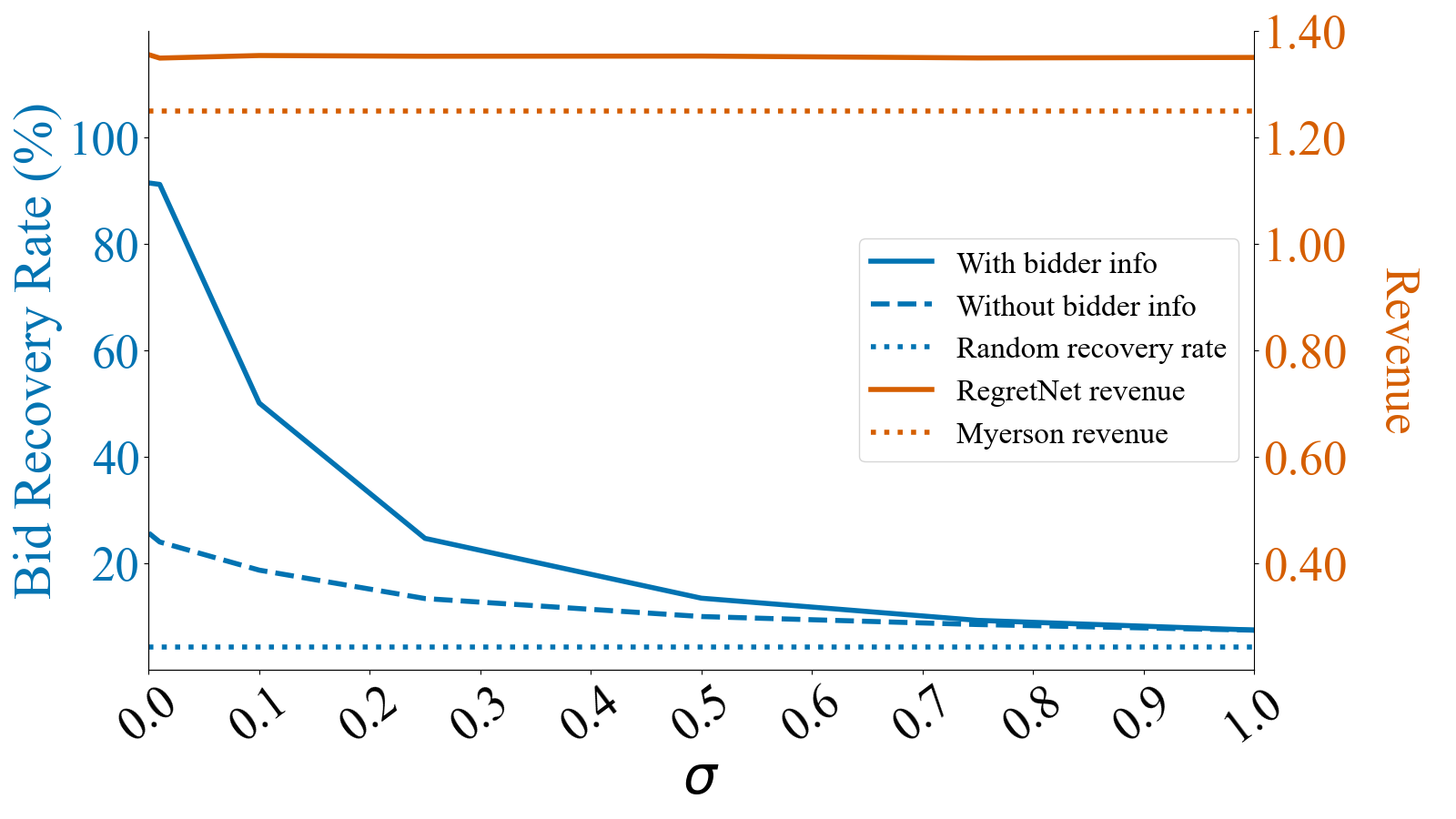}
    \vspace{-22pt}
    \caption{\textbf{Privacy and revenue in 2 $\times$ 3 auctions.} The recovery rate measures how often the inverted bids are within $\pm 0.02$ of the true values.}
    \label{fig:cost-2x3}
\end{minipage}
\hspace{8pt}
\begin{minipage}{\columnwidth}
    \centering
    \includegraphics[width=\textwidth]{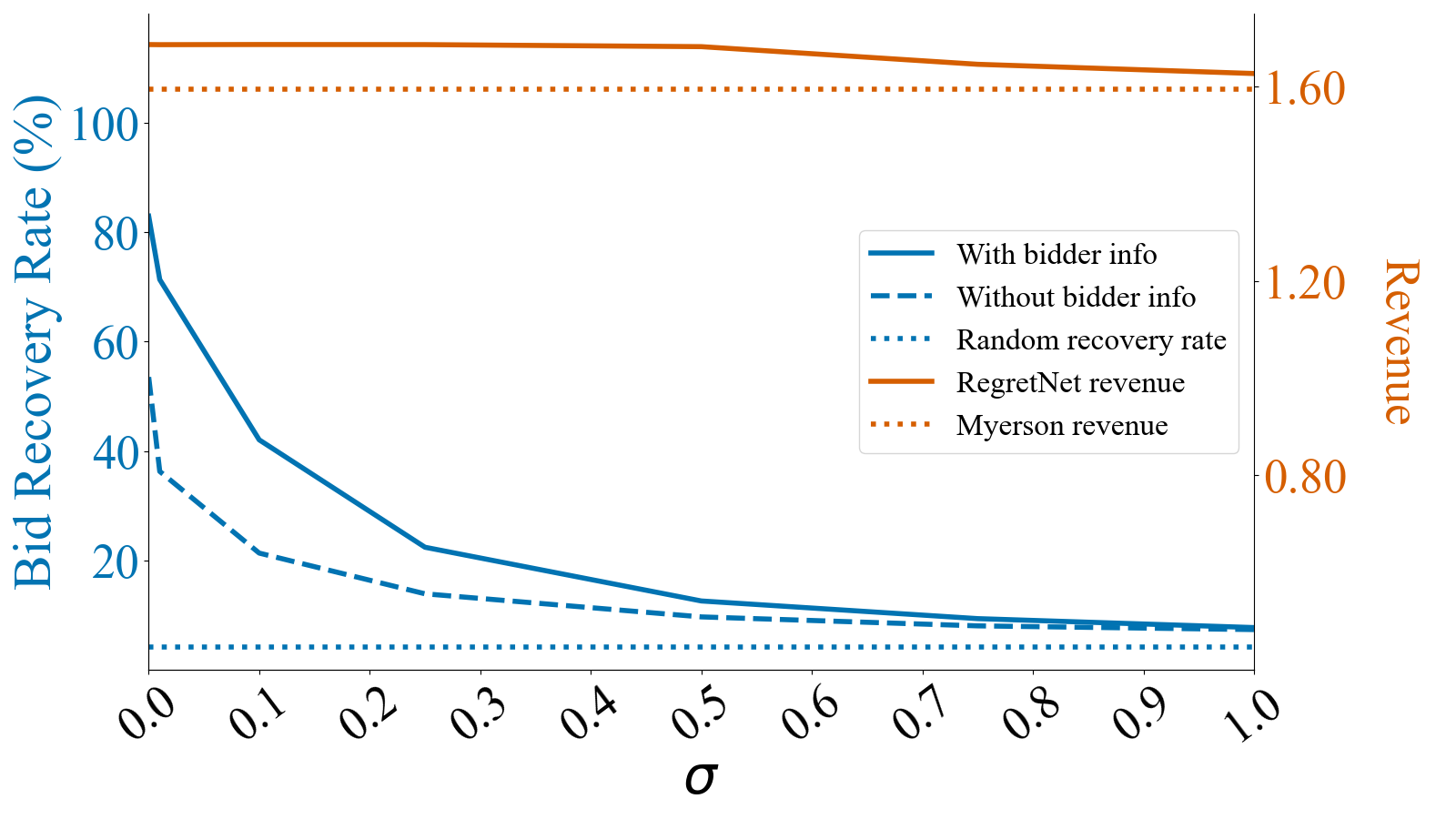}
    \vspace{-22pt}
    \caption{\textbf{Privacy and revenue in 3 $\times$ 3 auctions.} The recovery rate measures how often the inverted bids are within $\pm 0.02$ of the true values.}
    \label{fig:cost-3x3}
    \end{minipage}
\end{figure}
\begin{figure}[ht!]
    \centering
    \includegraphics[width=\columnwidth]{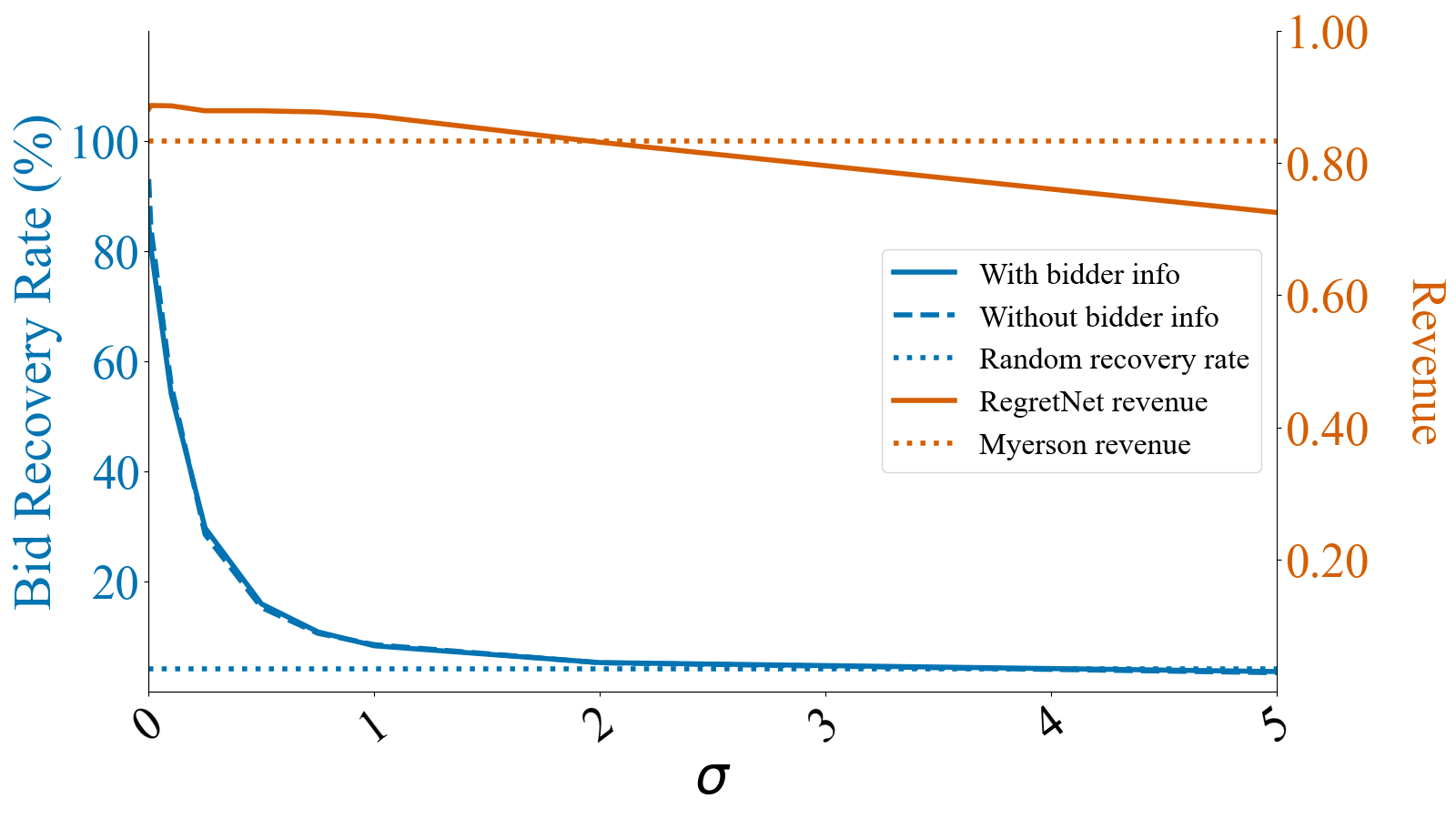}
    \includegraphics[width=\columnwidth]{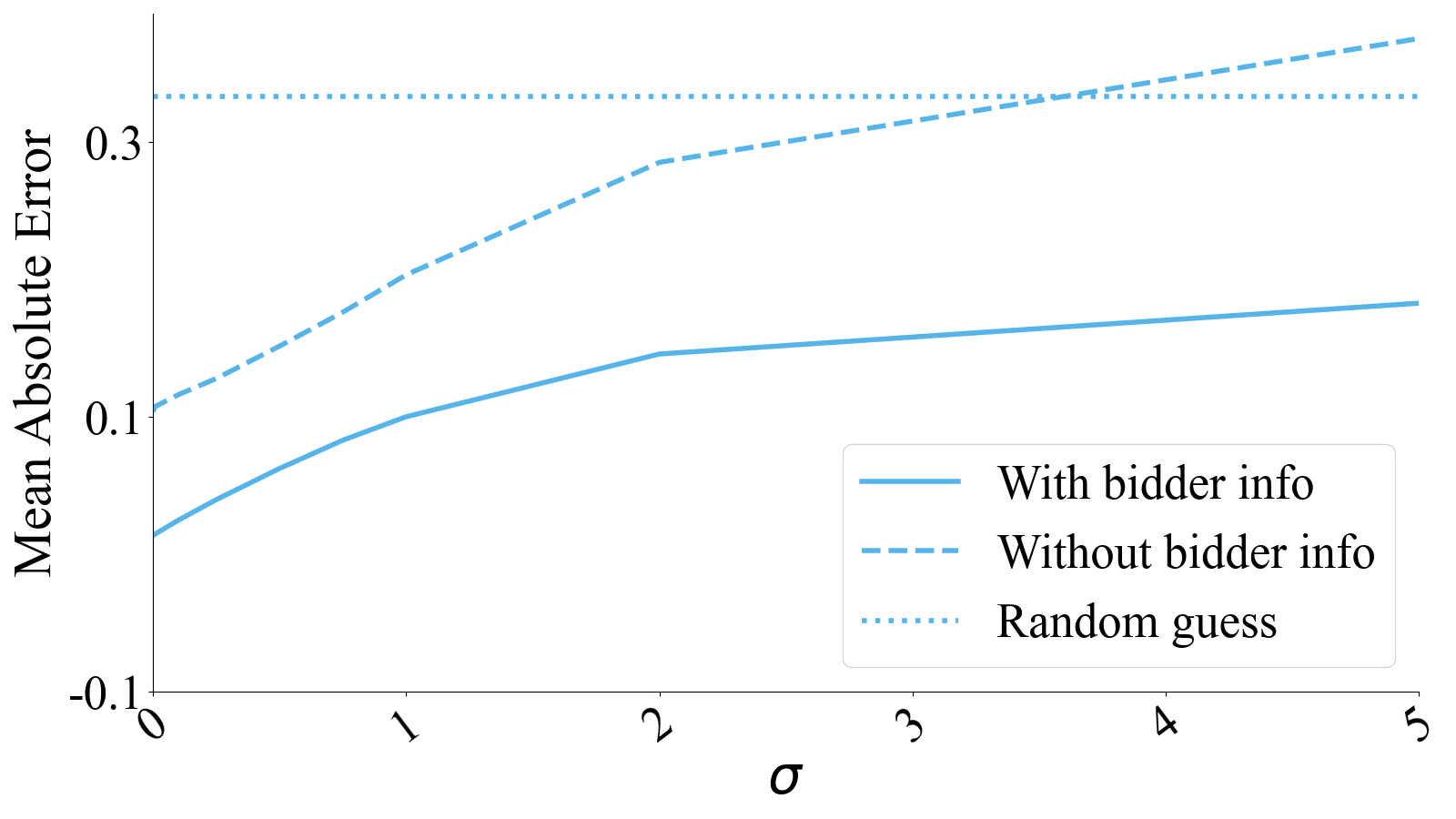}
    \vspace{-12pt}
    \caption{\textbf{Wide view of 2 $\times$ 2 auctions.} The recovery rate, the MAE, and the revenue for large values of $\sigma$ show that privacy and revenue decay quickly with $\sigma$.}
    \label{fig:cost-2x2-zoom}
\end{figure}

We show similar trends in the $3\times2$ and $2\times3$ auctions, whose privacy-revenue trade-offs are shown in Figures \ref{fig:cost-2x3} and \ref{fig:cost-3x3}. From these results, we identify two major take-aways. First, the cost to the auction house to reduce the recovery rates to 10\% is less than 1\% of the average revenue from unperturbed auctions; and these final reduced revenues are all above the Myerson threshold. Second, the scales on the left-hand vertical axes of Figures~\ref{fig:cost-2x3} and~\ref{fig:cost-3x3} indicate that even with more items these mechanisms leak private information.

An auction house that requires complete privacy may be interested in finding a $\sigma$ value large enough that the adversary is not able to extract more information than a random guess. For example, for $2\times2$ auctions, Figure \ref{fig:cost-2x2-zoom} shows that the recovery rate decays exponentially but the revenue can fall below the Myerson revenue. With this in mind, we show that for total privacy, $\sigma$ must be large enough that the revenue falls by around 20\%. In other words, there is a huge cost for complete privacy in these mechanisms. See Appendix \ref{sec:app-extended-results} for similar plots from other sizes of auctions.

\section{Discussion}
\label{sec:discussion}

We illuminate a privacy vulnerability of learned auction mechanisms that warrants attention.
Auction houses are interested in offering their bidders privacy and we show that with neural networks this comes at a cost. 
We propose a defense for model inversion attacks that can introduce privacy to already-trained models. Lastly, we analyze how much revenue the auction house can expect to lose for various degrees of privacy and we find that total privacy comes at an extremely high cost.

We believe that privacy in auctions is a critical consideration when evaluating the efficacy of the mechanism.  As such, neural auctions lack of privacy should be considered a significant flaw and is worth exploring further.  In particular, there might be other inversion methods that would pose an even greater threat to the privacy of neural auctions.  Additionally, While Differential Privacy is outside the scope of this paper given that it is primarily concerned with learning information about training data rather than the mechanism, auctions trained on real data would have the additional requirement of safeguarding training datasets.


We see our approach as a proof of concept that neural auctions are not private.  
To our knowledge, neural auction mechanisms have not yet been deployed in practice, so we do not think that our proof-of-concept attacks pose an immediate risk to privacy.
We view thinking through the privacy of neural auctions as being of interest in itself, and important before they are deployed.
Even in the face of a very naive attack, it is quite easy to infer information about bids by looking at the allocations. 
We present a method that can prevent this; however, our current model of an attacker is relatively naive and it is possible that a craftier adversary may be successful yet. We do not give any formal guarantees in the spirit of differential privacy, although this might be a fruitful direction for future work (for example, by bounding the Lipschitz constants of the auction networks~\citep{anil2019sorting,yoshida2017spectral}). 
Future work might also wish to consider larger auctions, and investigate the privacy properties of other neural auction architectures besides RegretNet 
(e.g., \citet{duan2022context, ivanov2022optimal, Rahme2021Auction}).

\bibliography{main}
\bibliographystyle{icml2023}

\newpage
\appendix
\onecolumn
\section{Appendix}

\subsection{Training}
\label{sec:app-training}

In our experiments, we use MLPs to model $1\times2$, $2\times2$, $3\times2$, $2\times 3$, and $3\times3$ approximately optimal auctions, the details of which we provide here.  Note that the training algorithm is a RegretNet implementation as proposed by~\citet{dutting2019optimal}.

All of our models have ReLU activations. Additionally, the allocation network has a softmax to ensure that each item's allocations sum to one, and a sigmoid activation function on the payment network so that the payments become a fraction between zero and one (to enforce IR). Models are trained on 640,000 sample valuation profiles $b_i\sim U[0,1]$ (split into batches of 128) using the Adam optimizer trained with an initial learning rate of $1e^{-3}$ for several epochs.

The training process includes adding an additional term for regret minimization (similar to the augmented Lagrangian method used in~\citep{dutting2019optimal}).  The corresponding weight for this regret term is used to update the regret in the loss periodically and the value of $\rho$ itself is also updated periodically. Finally, the agents' ex post regret is optimized in an inner loop, whose parameters are the last three rows of Table \ref{tab:app-hyperparams}. See the original RegretNet paper for more details and the derivation of the loss function \citep{dutting2019optimal}.

\begin{table}[ht!]
\centering
\caption{\textbf{Hyperparameters.} We provide training hyperparameters for all auction sizes we study.}
\label{tab:app-hyperparams}
\begin{tabular}{lccccc}
\toprule
Auction Size                          & 1$\times$2   & 2$\times$2   & 3$\times$2   & 2$\times$3   & 3$\times$3   \\
\midrule
Epochs                                & 10    & 30    & 20    & 20    & 30    \\
Train Batch Size                      & 128   & 128   & 128   & 128   & 128   \\
Initial Learning Rate                 & 0.001 & 0.001 & 0.001 & 0.001 & 0.001 \\
Number of Hidden Layers               & 3     & 3     & 5     & 5     & 5     \\
Hidden Layer size                     & 100   & 100   & 100   & 100   & 100    \\
$\rho$ Update Period (Epochs)         & 2     & 2     & 2     & 2     & 8     \\
Lagrange Weight Update Period (Iters) & 100   & 100   & 100   & 100   & 100   \\
Initial $\rho$                        & 1     & 1     & 1     & 1     & 0     \\
$\rho$ Increment                      & 10    & 5     & 1     & 1     & 1     \\
Initial Lagrange Weight               & 5     & 5     & 5     & 5     & 5     \\
Misreport Learning Rate (Training)    & 0.1   & 0.1   & 0.1   & 0.1   & 0.1   \\
Misreport Iterations (Training)       & 25    & 25    & 25    & 25    & 25    \\
Misreport Initializations (Training)  & 10    & 10    & 10    & 10    & 10   \\
\bottomrule
\end{tabular}
\end{table}

\subsection{Model Inversion}
\label{sec:app-inversion}

The adversary is able to invert the auctions by means of a white box attack.  In this setting, the adversary has access to the outputs $(g, p)$ of the neural auctions as well as the weights of the mechanisms. They are able to query the mechanism with bids, observing the outputs of the auction each time. Using a set of simulated bids, output pairs ($b \rightarrow (g,p)$) the adversary is able to reverse the process, using true payments and allocations and recovering the corresponding bids: $(g,p)\rightarrow b$. 

The inversion algorithm is projected gradient descent using the Adam optimizer. It iteratively updates its guessed bids, such that the difference between the resulting payments and allocations and the true payments and allocations is minimized. This algorithm relies on access to the trained neural auction model so that the guessed bids can move in the direction of the corresponding gradients that minimize the inversion objective. We run the inversion with an initial learning rate of $0.002$ for 50,000 iterations. We compare the recovered bid to the true bid to within a tolerance of $0.02$ to measure success.

\subsection{Privacy in Myerson Auctions}
\label{sec:myerson}

In order to understand the privacy ramifications of employing neural auctions we establish baselines with Myerson auctions. For the valuation distributions we deal with, the Myerson auction is simply a second price auction with reserve. A single item with a reserve price $r$ is sold for a payment $p = \max(b_2, r)$, where $b_2$ is the second highest price. If this price causes the auction to violate IR (if $b_i<r, \forall i$), then the auctioneer would prefer not to sell the item and the payment would be zero. Like neural auctions, Myerson auctions leak some information that can help an adversarial observer, (or active bidder), guess the bids of the market participants.  The adversary's inversion accuracy ($\mathcal{A}$) depends on the on their guessing strategy and the information they have.

\subsubsection{Naive Guessing}
An adversary may deploy a naive guessing strategy by observing the public payments and allocations and naively guessing that each bidder bid the payment amount. Such a bidding strategy would produce a lower bound on $\mathcal{A}$.
\begin{lemma} By employing a naive strategy in which the adversary guesses that all bidders bid the payment amount, their accuracy is bound from below as follows.
    $$\mathcal{A} \geq \left(1-\left(r^n + (r)^{n-1}(1-r)\right)\right)*\Bigg(\frac{1}{n}\Bigg)$$
\label{eq:myerson-bound}
\end{lemma}
\vspace{-24pt}
\begin{proof}
We can break down the expected guessability of the auction using this naive strategy by analyzing the three separate cases.  (I) All bids are less than the reserve price, or $b_i<r\, \forall i$. This occurs with probability $r^n$, which results in no sale and limits the adversary to randomly guessing all bids.  (II) Only one of the bids is higher than the reserve price, which implies $p=r$.  This case has a probability of $n(r)^{n-1}(1-r)$ and the adversary will only learn the reserve price (which need not be equivalent to any of the bids). (III) At least two bids are above the reserve price which occurs with probability $1-(r^n + n(r)^{n-1}(1-r))$.  In this case the adversary is guaranteed to be right for at least $\frac{1}{n}$ bids, since the payment amount is exactly the second highest bid.  This produces the desired lower bound on the adversary's inversion accuracy $\mathcal{A}$.
\end{proof}

\subsubsection{Intelligent Guessing}
We can improve the inversion strategy with additional information gleaned from the payments and allocations for cases (I) and (II).  If the auction has no winner, as in case (I), the adversary knows that all bids fall below the reserve price so they can adjust their distribution to assume that $b_i\sim U[0,r]$.  Similarly, in case (II), when the payment price is equal to the reserve price, the adversary can deduce that only one bid lies above the reserve price and the rest lie below. This improves the adversary's ability to guess bids when in cases (I) and (II).

\subsubsection{Adversarial Bidder}
Another threat model is that of an active adversarial bidder that can use their own bid as an additional piece of information to improve their guessing accuracy.  If the bidder wins the auction, they learn no additional information about the other bidders since they only know that everyone else bid at or below the payment, information known to an external adversary as well.  However, if the adversary observes that the payment is equal to their bid, they know that they set the clearing price of the auction and that every bidder, except the winner, bid at or below them.  This private information, coupled with the improved guessing strategy for when the reserve price affects the auction outcome make the auction more invertible and therefore less private.  These results are detailed in Table~\ref{tab:myerson-recovery-rates}. Note, the empirical success rates in Table \ref{tab:myerson-recovery-rates} are all above the corresponding bounds computed with the formula in Equation \eqref{eq:myerson-bound}.

\begin{table}[ht!]
    \centering
    \caption{\textbf{Myerson auction bid recovery rates.} Empirical measurements of the adversary's success rate (\%) for multiple guessing strategies in an auction with a reserve price of $0.5$ for all items show that private information is leaked. Note that these numbers are from empirical experiments where predictions that recover bids within $\pm0.02$ of the true value are considered successful.}
    \label{tab:myerson-recovery-rates}
    \vspace{8pt}
    \small
    \begin{tabular}{lcccccccc}
    \toprule
                                        & \multicolumn{3}{c}{Bidders $\times$ Items} \\
                                        & 2 $\times$ 2   & 3 $\times$ 2  & 3 $\times$ 3\\
    \midrule
        Naive  (no bidder info)        & $16.493$       & $20.655$      & $20.580$ \\
        Intelligent (no bidder info)   & $19.234$       & $22.831$      & $22.775$ \\
        Intelligent (with bidder info) & $19.219$       & $23.519$      & $23.520$ \\
        \bottomrule
    \end{tabular}
\end{table}

\subsection{Extended Results}
\label{sec:app-extended-results}

\begin{figure}[ht!]
    \centering
    \includegraphics[width=0.49\textwidth]{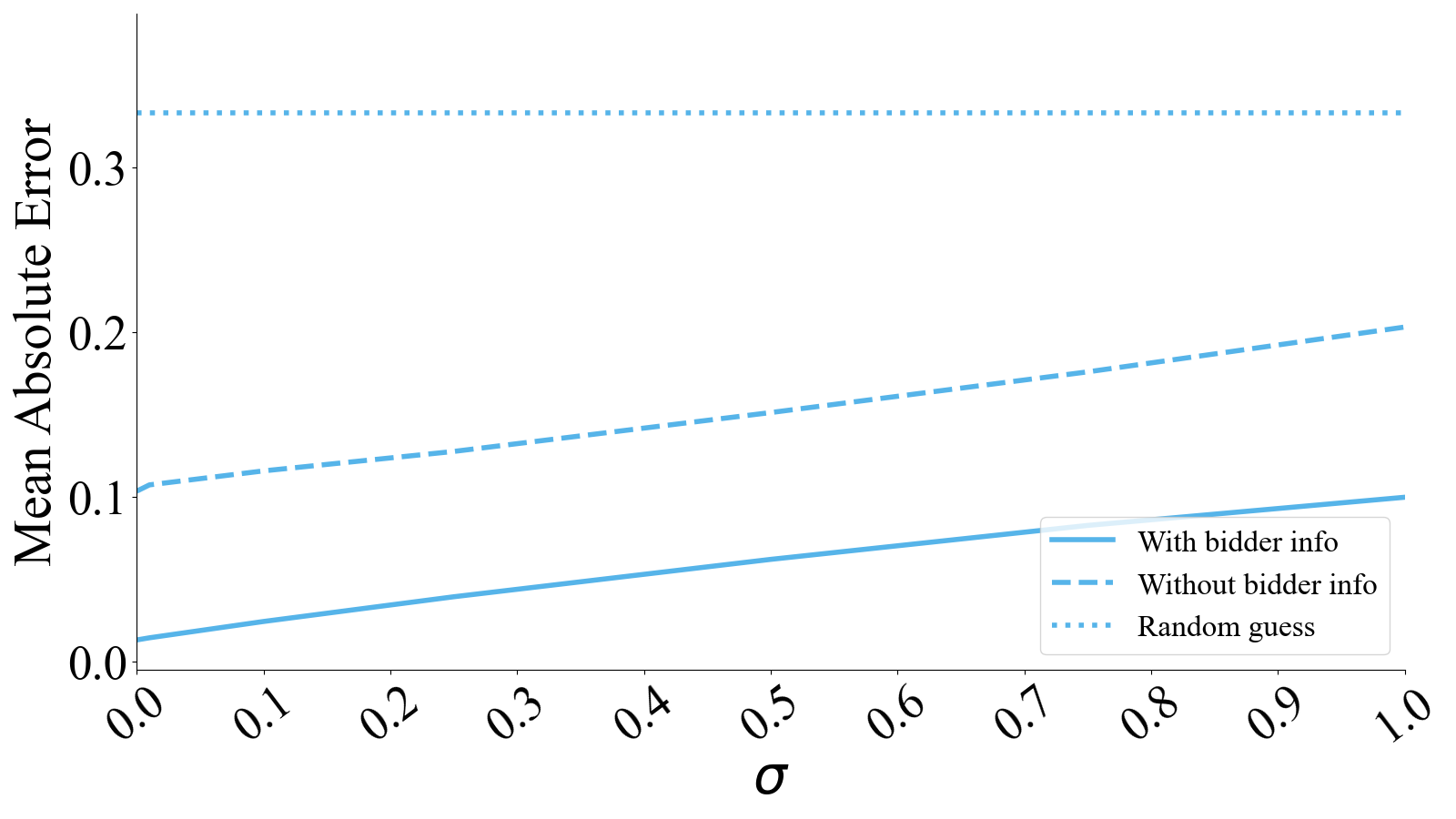}   \includegraphics[width=0.49\textwidth]{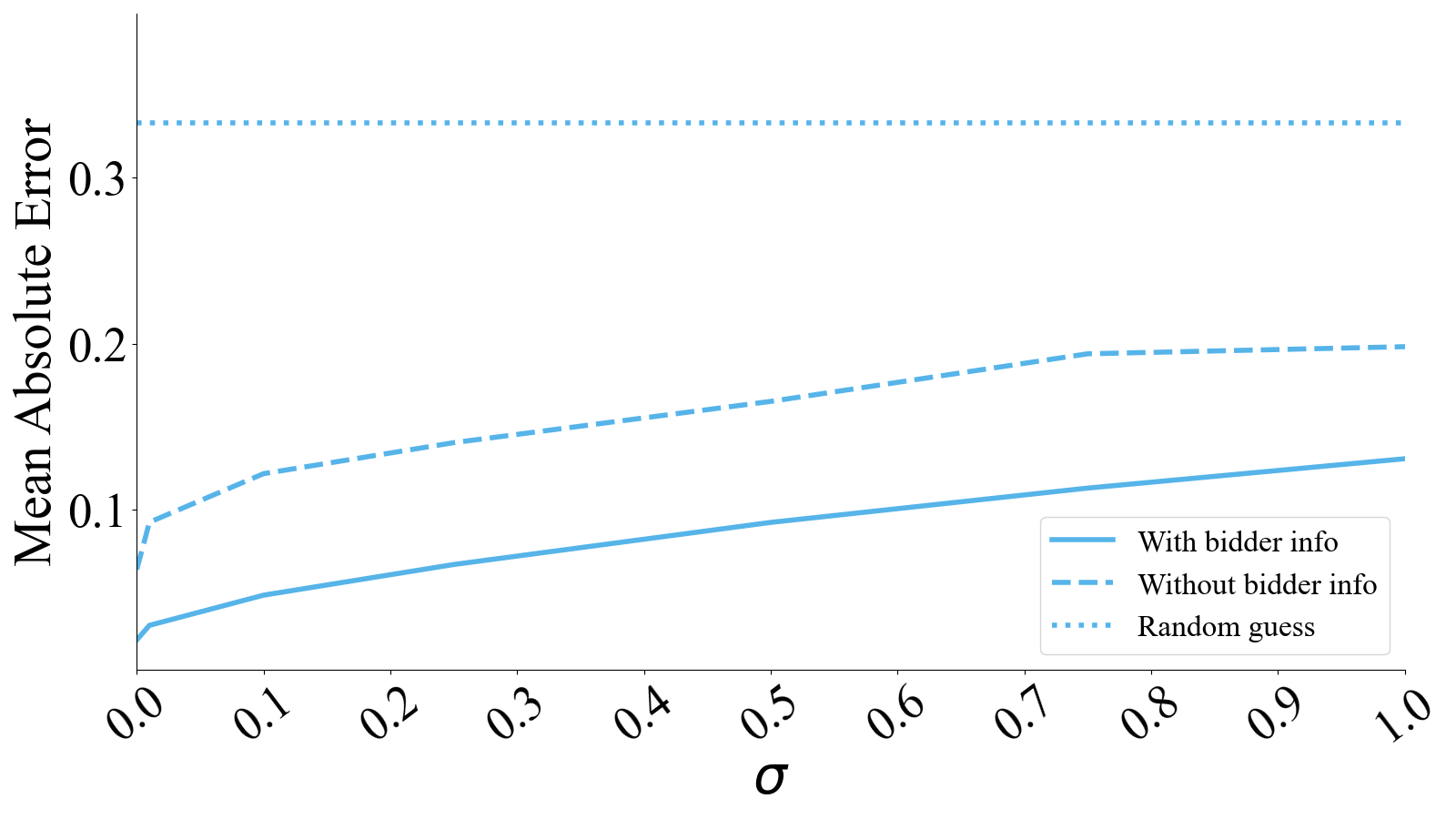}
    \caption{\textbf{MAE for $2\times3$(left) and $3\times3$ (right) auctions.} For values of $\sigma$ shown in Figures \ref{fig:cost-2x3} and \ref{fig:cost-3x3}, we show the corresponding MAE. With larger magnitude noise, the adversary's error rises.}
    \label{fig:app-cost-2x3-and-3x3-zoom}
\end{figure}

\begin{figure}[ht!]
    \centering
    \includegraphics[width=0.49\textwidth]{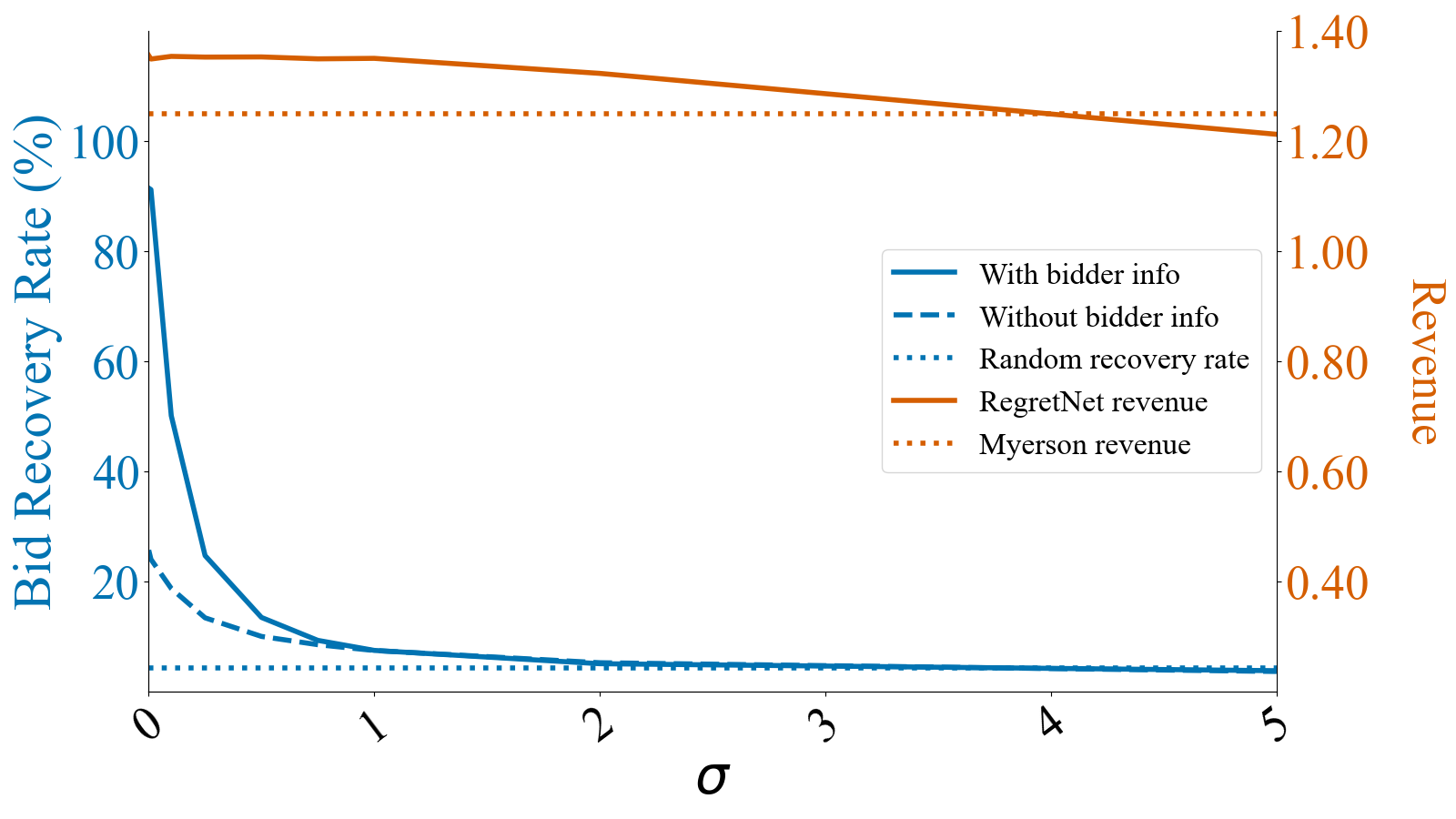}
    \includegraphics[width=0.49\textwidth]{figures/2x3-mae-zoomed.png}   
    \caption{\textbf{Wide view of 2 $\times$ 3 auctions.} The recovery rate, the revenue, and the MAE for large values of $\sigma$ show that privacy and revenue decay with $\sigma$.}
    \label{fig:app-cost-2x3-zoom}
\end{figure}

\begin{figure}[ht!]
    \centering
    \includegraphics[width=0.49\textwidth]{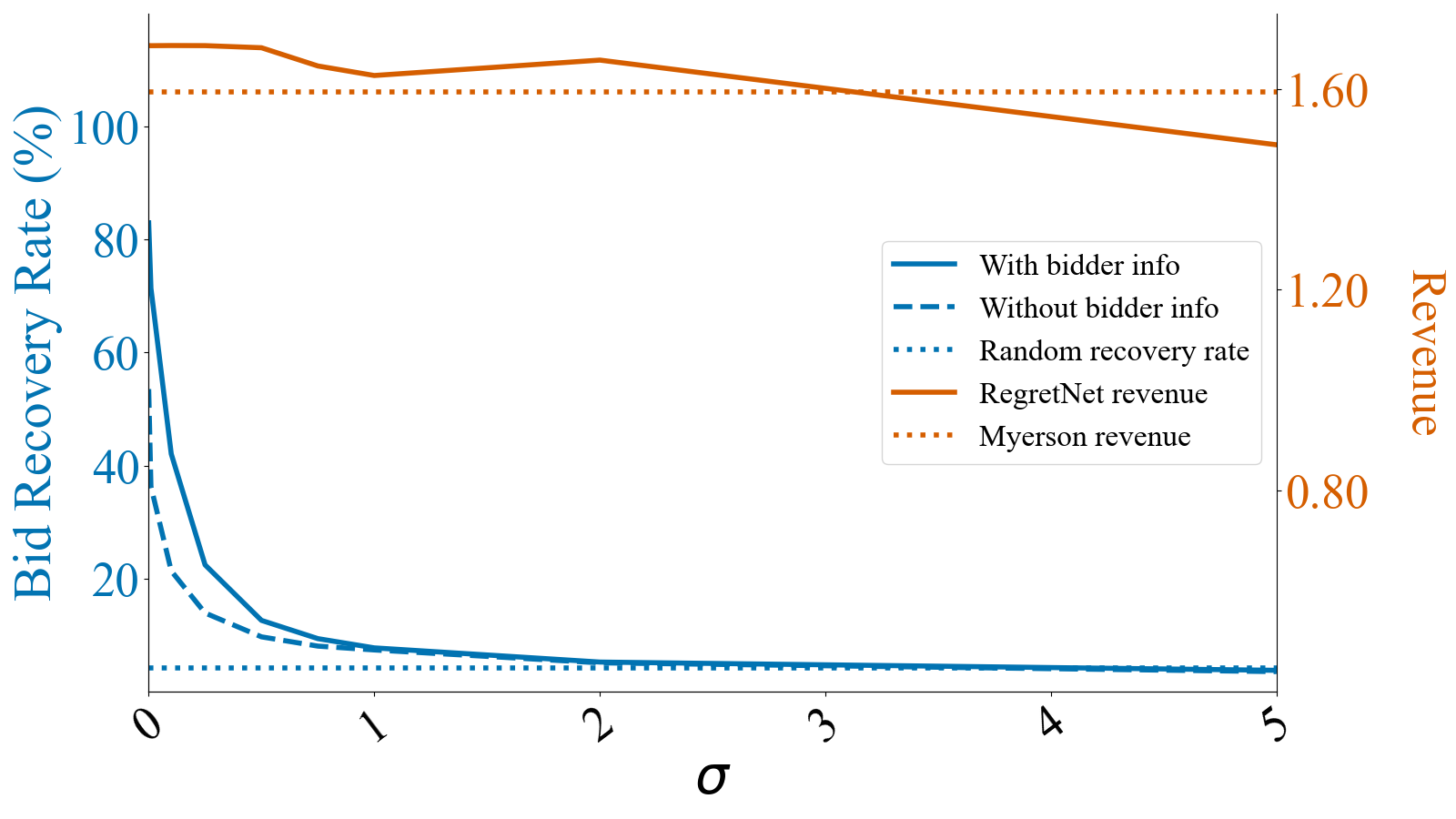}
    \includegraphics[width=0.49\textwidth]{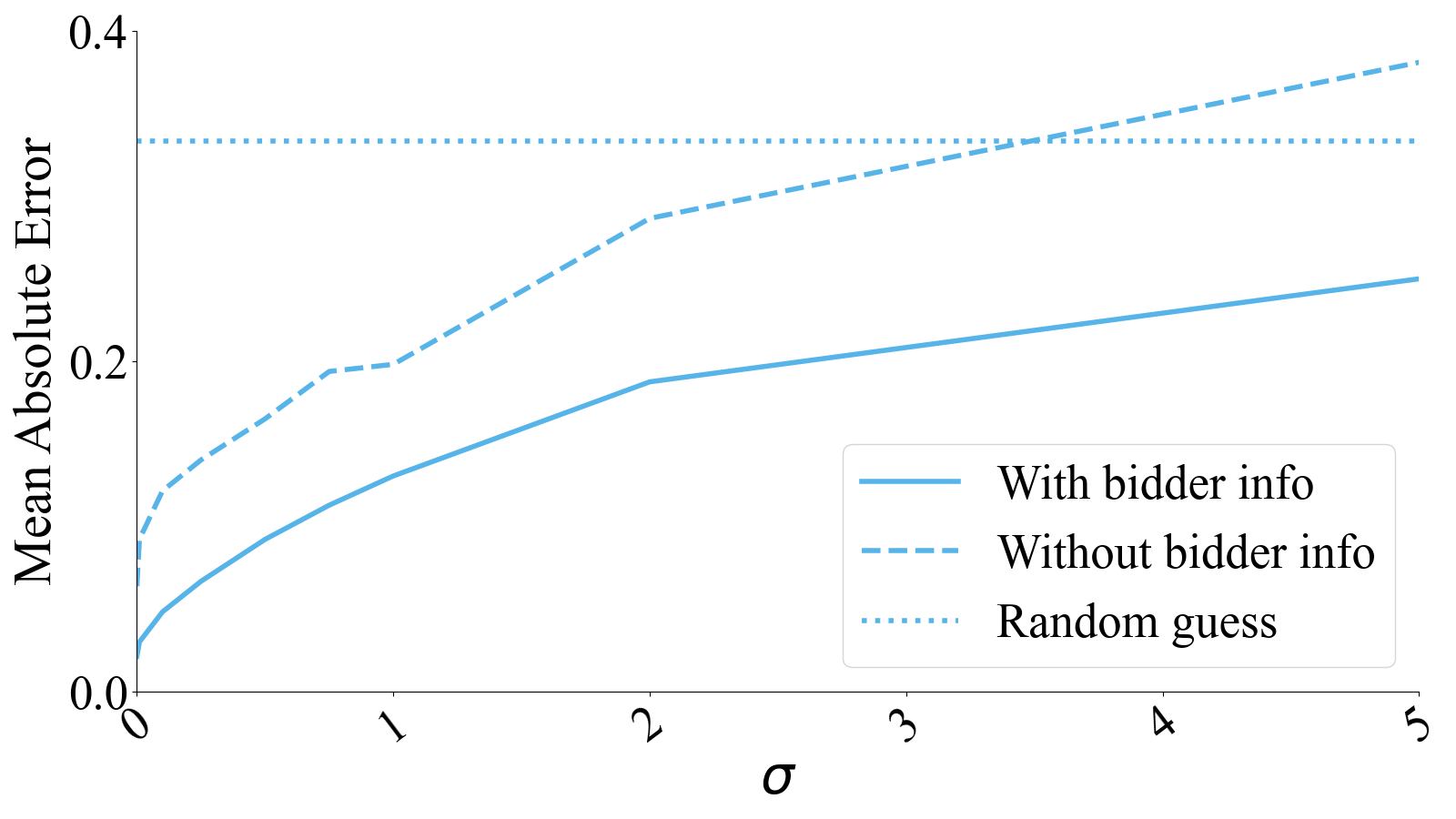}   
    \caption{\textbf{Wide view of 3 $\times$ 3 auctions.} The recovery rate, the revenue, and the MAE for large values of $\sigma$ show that privacy and revenue decay with $\sigma$.}
    \label{fig:app-cost-3x3-zoom}
\end{figure}

Figure~\ref{fig:app-cost-2x3-and-3x3-zoom} shows the MAE for the inversion of the $2 \times 3$ and $3 \times 3$ auction models. These results compliment Figures~\ref{fig:cost-2x2} and \ref{fig:cost-3x3} in the main body. Furthermore, Figures~\ref{fig:app-cost-2x3-zoom} and \ref{fig:app-cost-3x3-zoom} show even more detail on the $2 \times 3$ and $3 \times 3$ auction inversion results.  As more noise gets added to the system (i.e. as $\sigma$ grows) the revenue decreases and and mean absolute error grows.  It is worth noting that with small values of $\sigma$, the auctions produce higher revenue than the Myerson auction while reducing bid recovery rate exponentially.

\begin{figure}[ht!]
    \centering
    \includegraphics[width=0.49\textwidth]{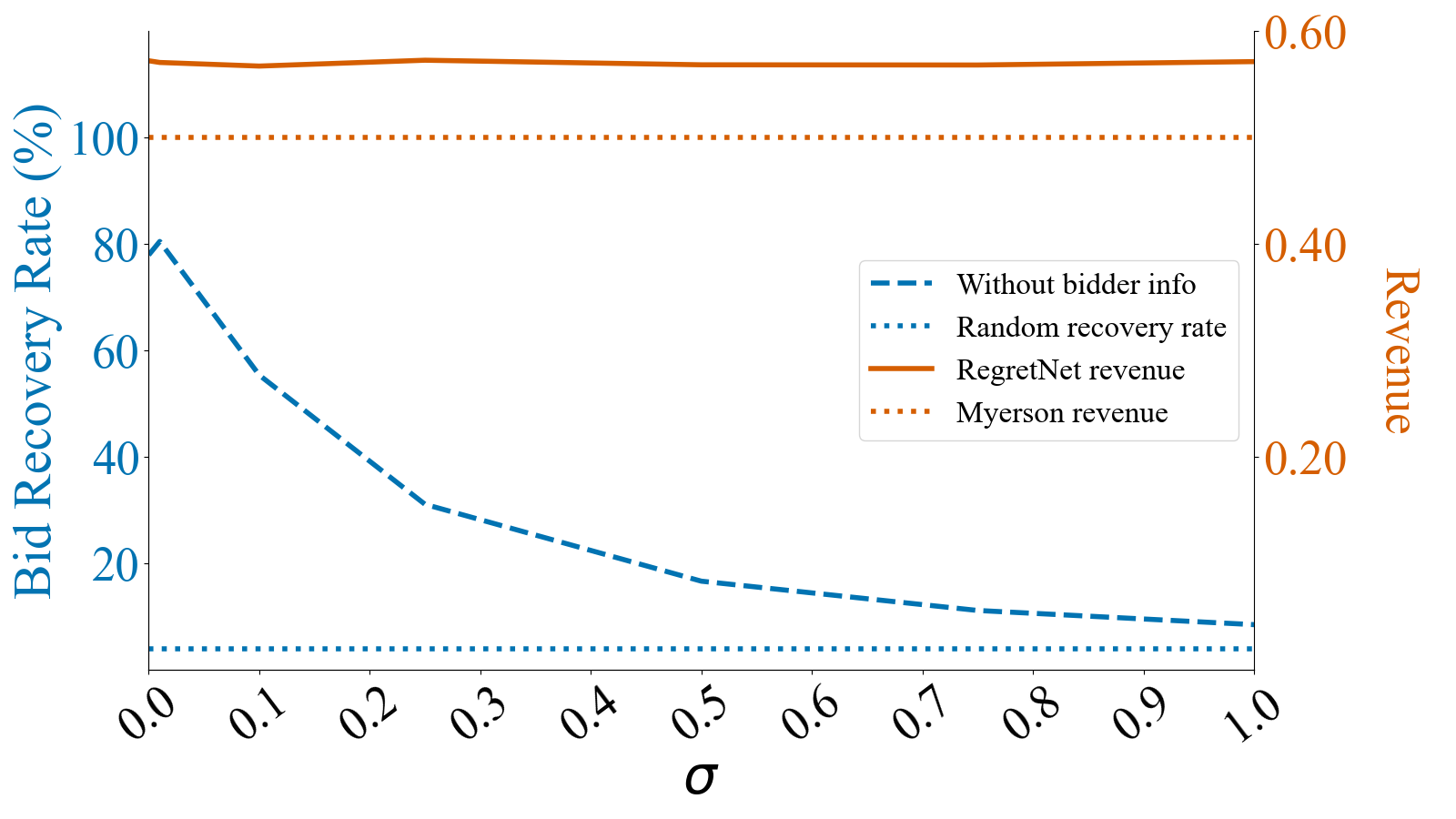}
    \includegraphics[width=0.49\textwidth]{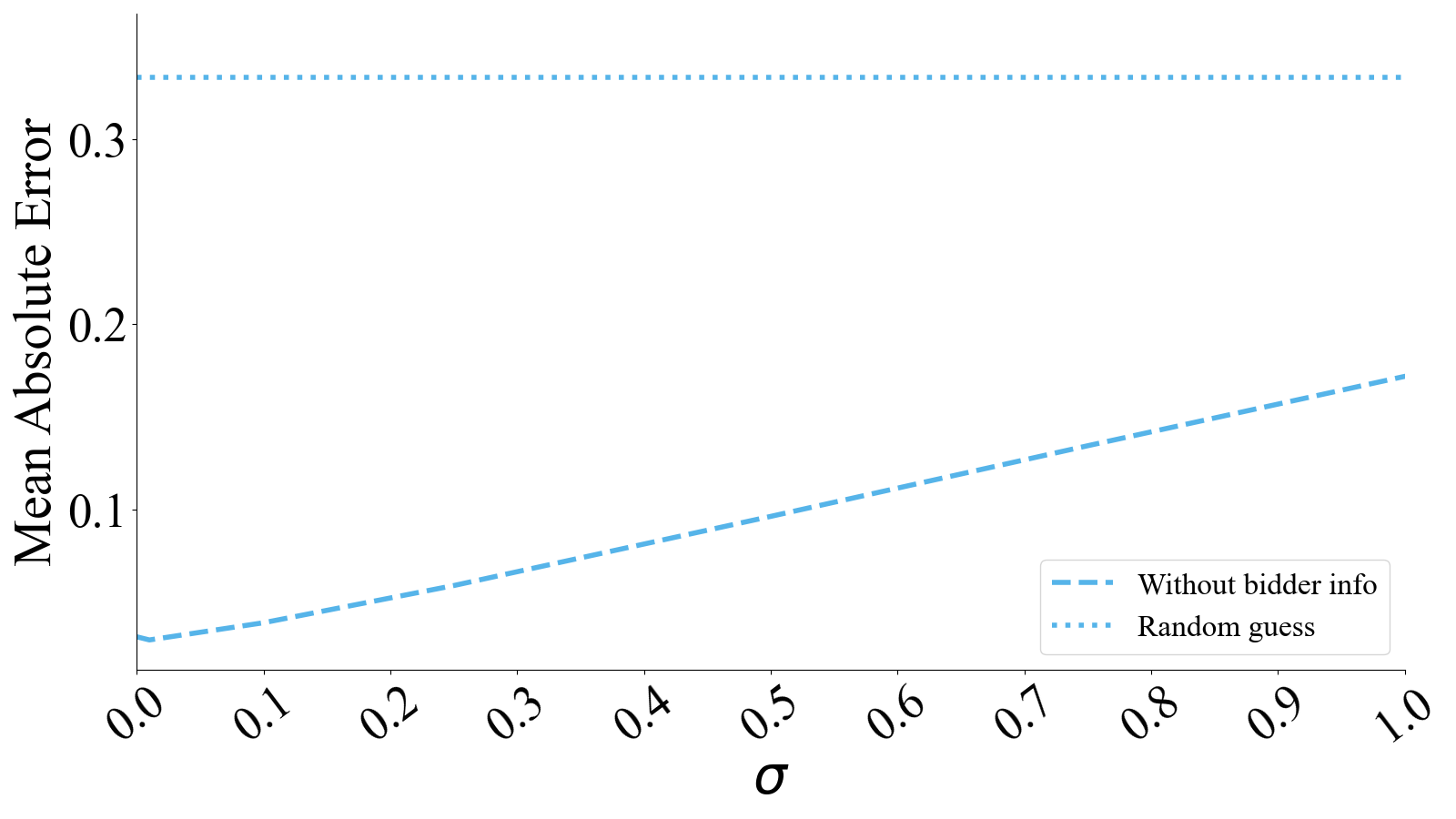}   
    \caption{\textbf{Privacy and revenue of 1 $\times$ 2 auctions.}}
    \label{fig:app-cost-1x2}
\end{figure}

\begin{figure}[ht!]
    \centering
    \includegraphics[width=0.49\textwidth]{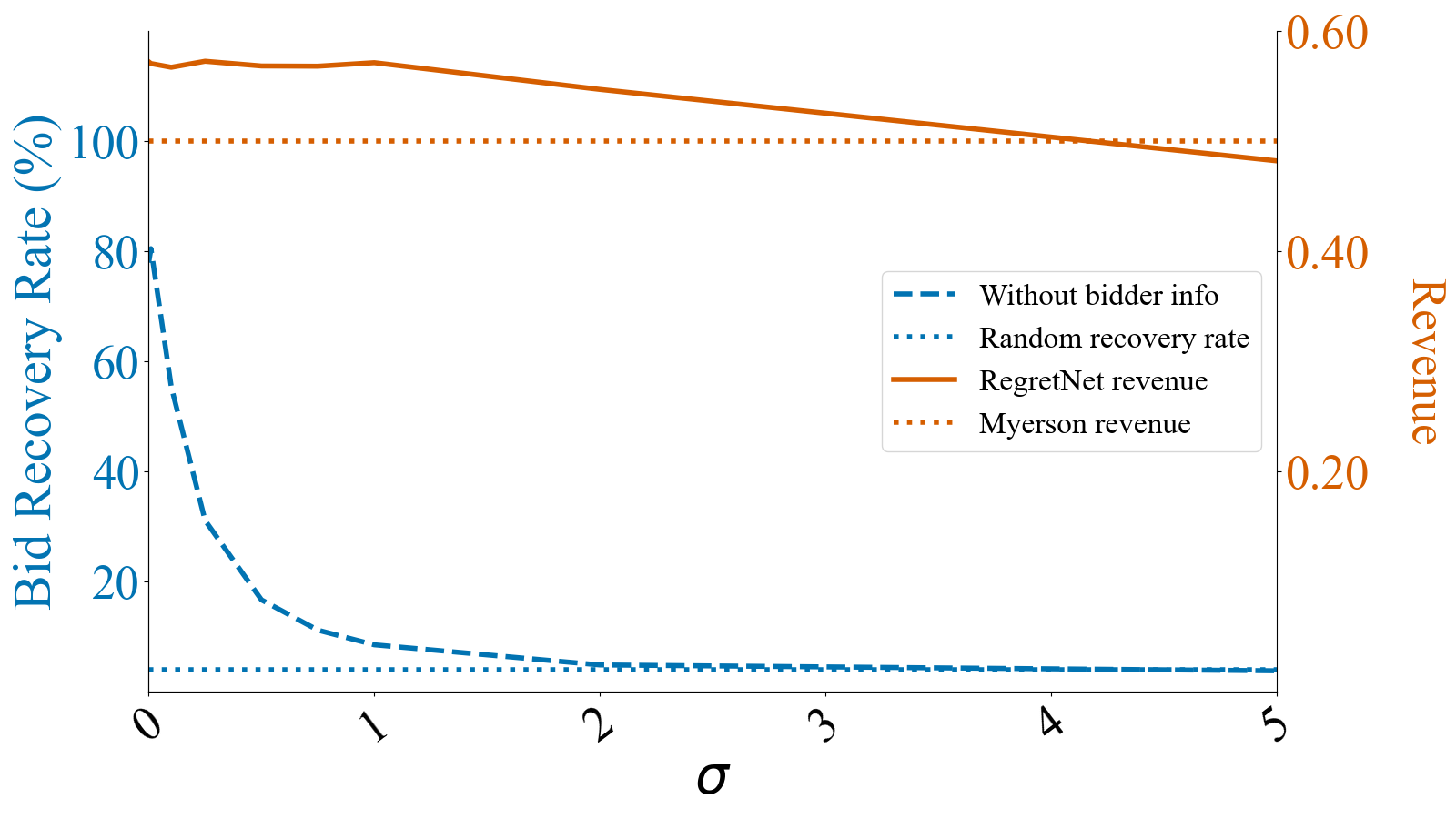}
    \includegraphics[width=0.49\textwidth]{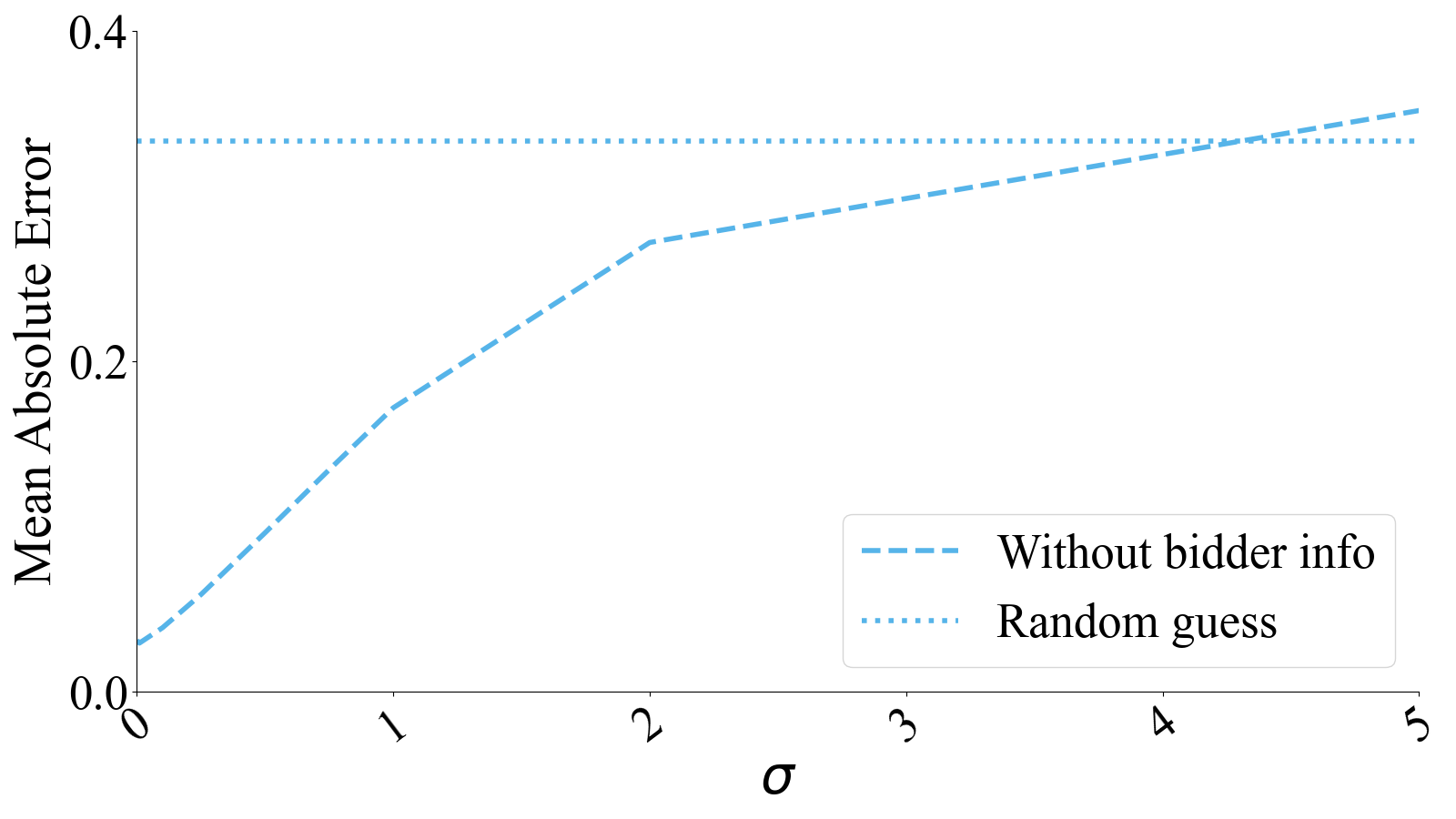}   
    \caption{\textbf{Wide view of 1 $\times$ 2 auctions.} The recovery rate, the revenue, and the MAE for large values of $\sigma$ show that privacy and revenue decay with $\sigma$.}
    \label{fig:app-cost-1x2-zoom}
\end{figure}

\begin{figure}[ht!]
    \centering
    \includegraphics[width=0.49\textwidth]{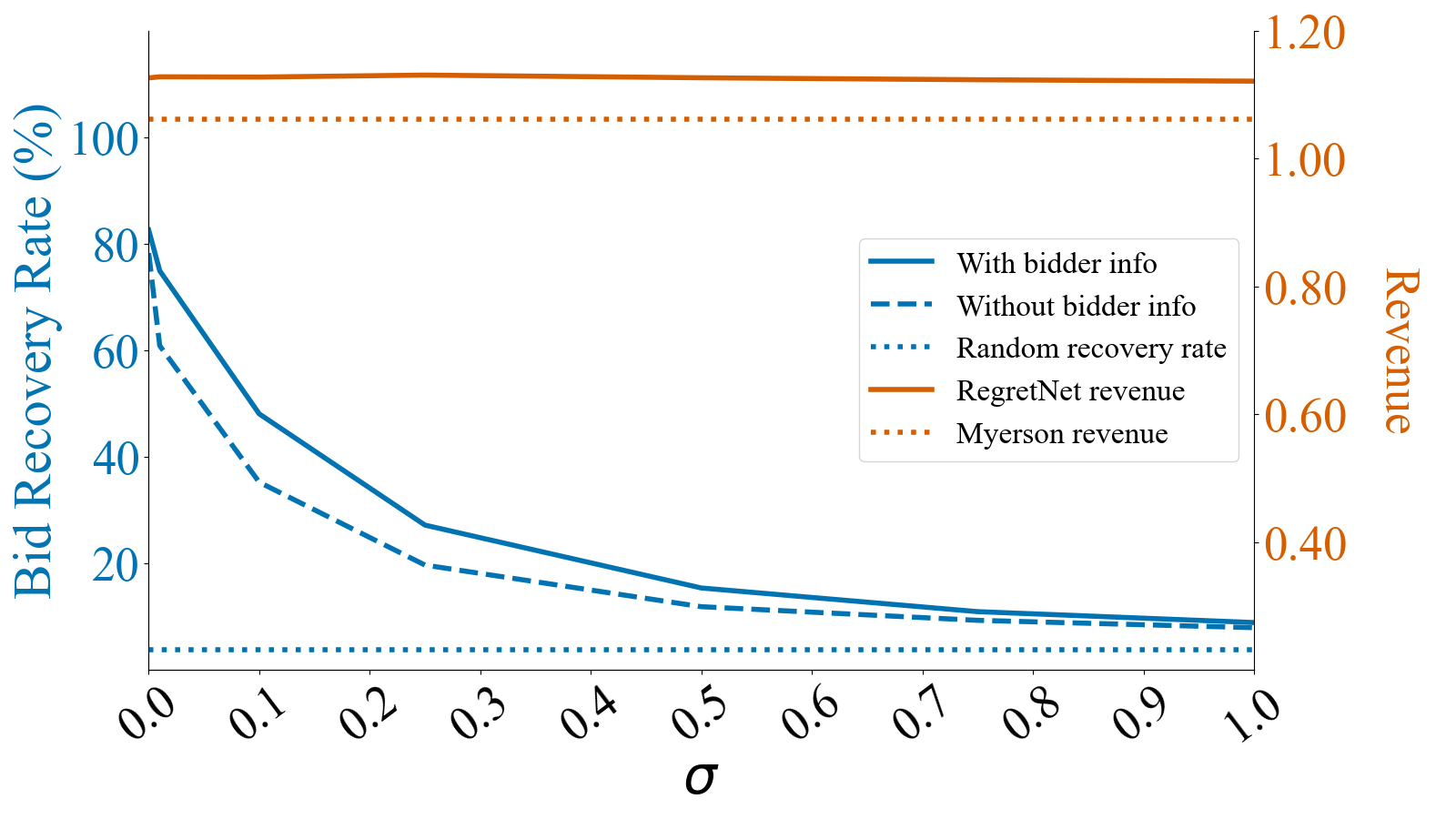}
    \includegraphics[width=0.49\textwidth]{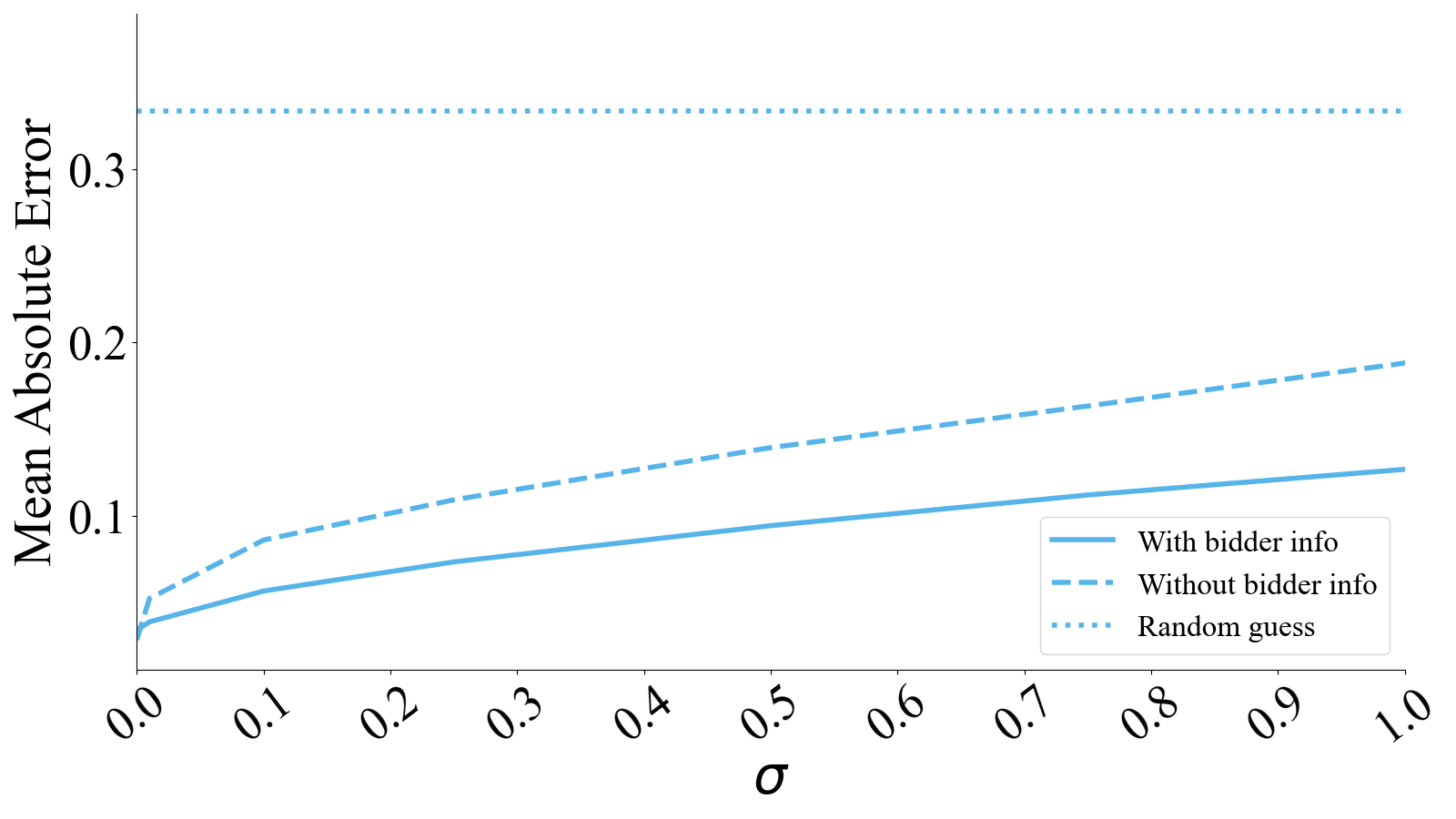}   
    \caption{\textbf{Privacy and revenue of 3 $\times$ 2 auctions.}}
    \label{fig:app-cost-3x2}
\end{figure}

\begin{figure}[ht!]
    \centering
    \includegraphics[width=0.49\textwidth]{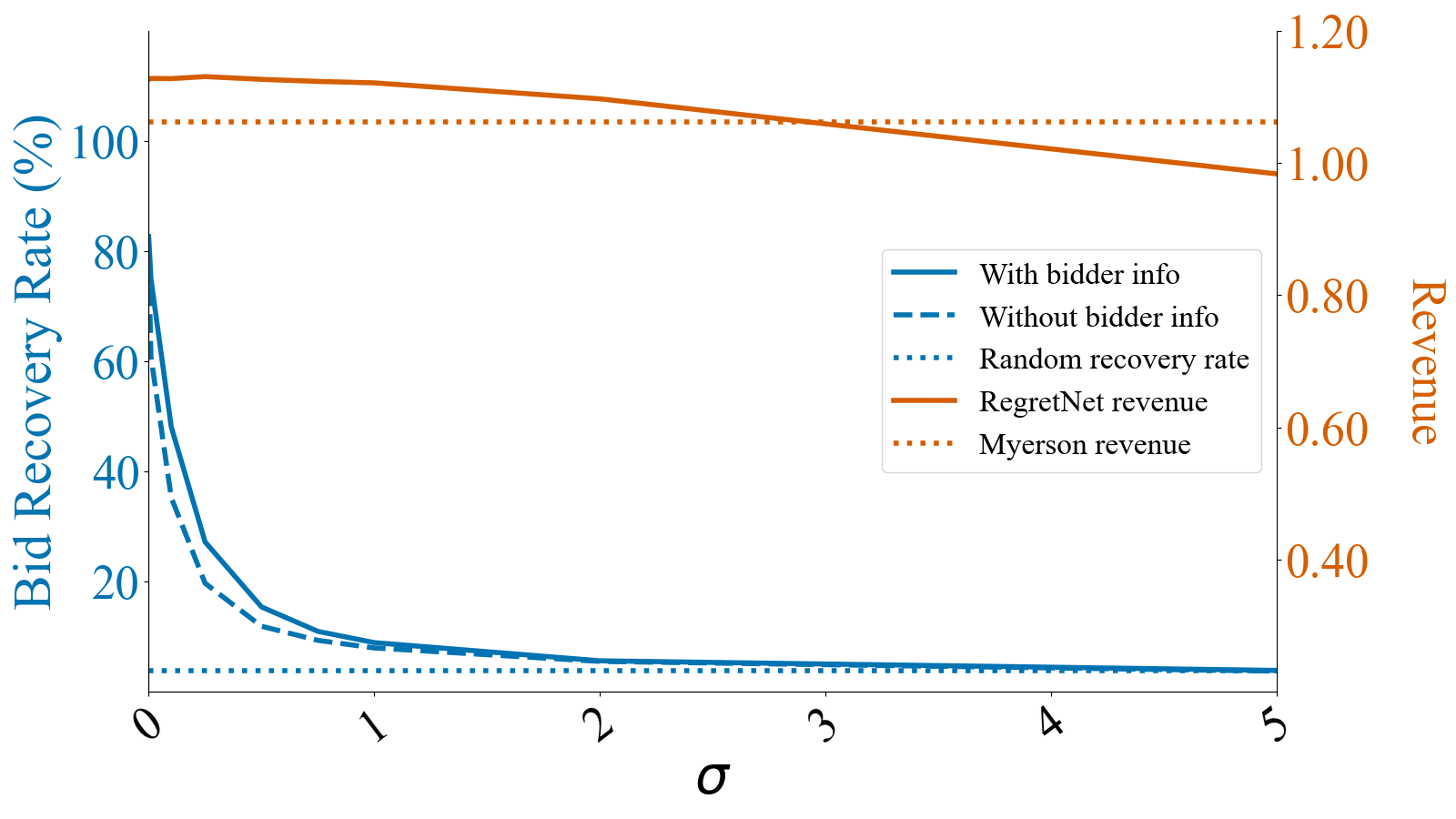}
    \includegraphics[width=0.49\textwidth]{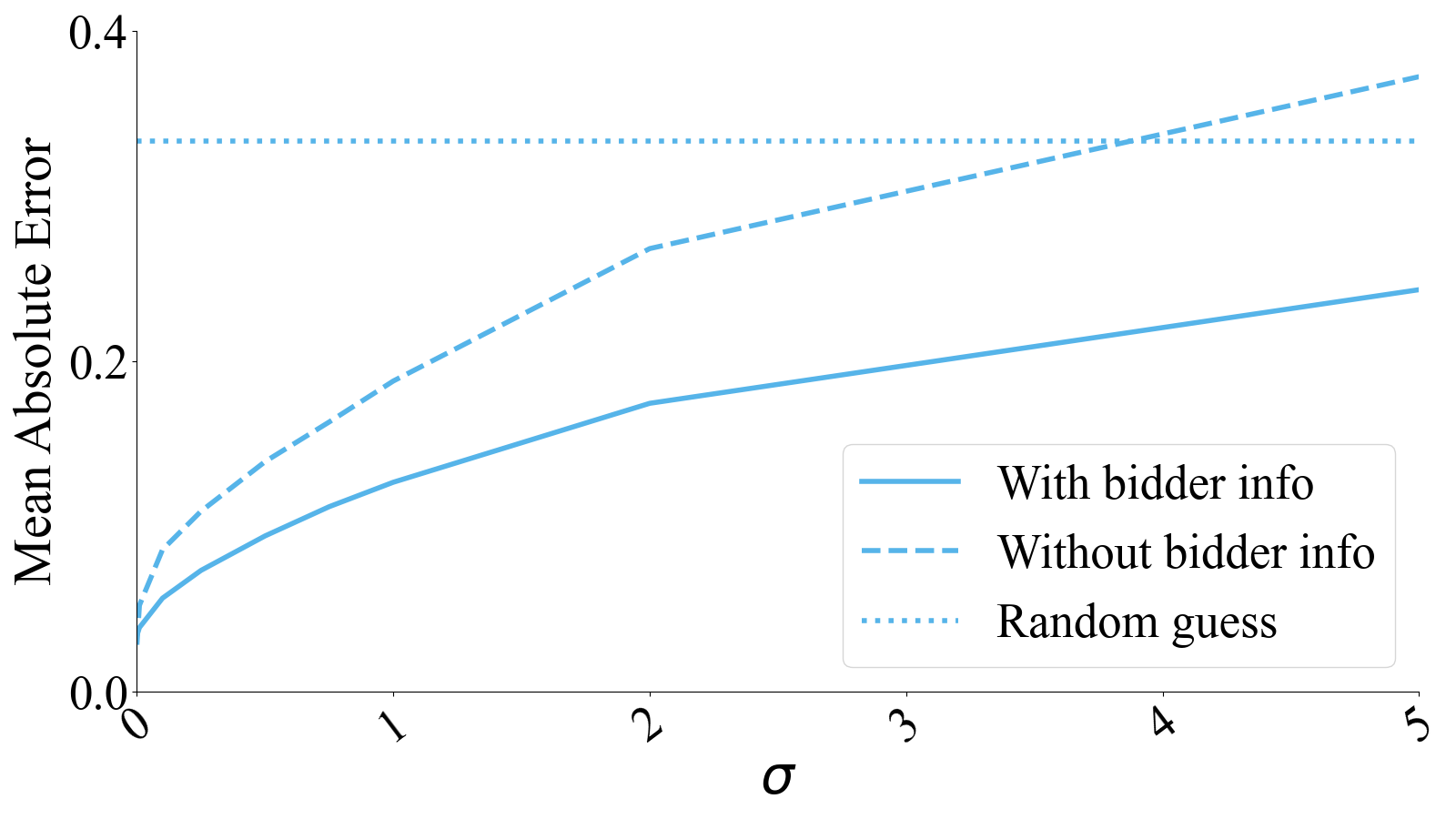}   
    \caption{\textbf{Wide view of 3 $\times$ 2 auctions.} The recovery rate, the revenue, and the MAE for large values of $\sigma$ show that privacy and revenue decay with $\sigma$.}
    \label{fig:app-cost-3x2-zoom}
\end{figure}

\begin{figure}[ht!]
    \centering
    \includegraphics[width=0.7\textwidth]{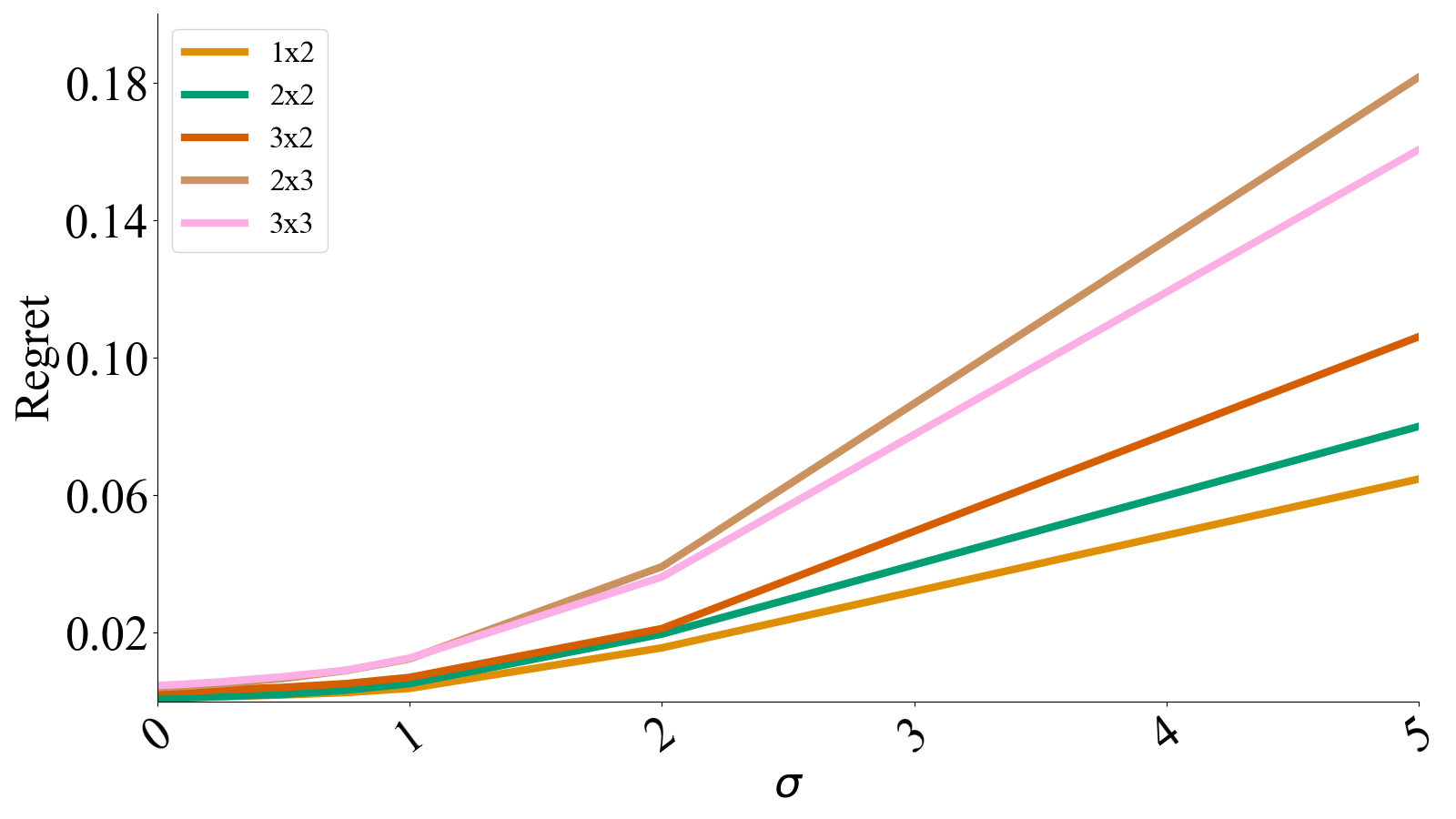}
    \caption{\textbf{Regret of noisy auctions.}}
    \label{fig:app-regret}
\end{figure}

Finally, Figures~\ref{fig:app-cost-1x2}-\ref{fig:app-cost-3x2-zoom} show the same results for the remaining auction sizes we consider. We also show how the regret changes with the magnitude of the noise in Figure \ref{fig:app-regret}.

\subsection{Adversaries with Limited Knowledge}
\label{sec:app-no-access}

While bidder valuation distribution is known in the general auction setting, one might wonder how well an adversary attempting to invert auctions can do without this knowledge. In the inversion strategy we propose, the initial guess is a random sample from the valuation distribution $P$. Additionally, we project the guess at each iteration onto the support of $P$ in order to enforce that our guess/inversion is feasible. If the adversary did not know $P$ exactly, they would not be able to sample from the actual distribution. Furthermore, if they do not even know the support of $P$ -- how well can they possibly do?

To address this question, we modify the inversion strategy as follows. We initialize all guesses to zeros and we only enforce that bids be positive -- a reasonable assumption for any adversary. Even without any information about $P$, these mechanisms are invertible. Table \ref{tab:app-recovery-rates-zeros} shows the inversion success rates when the adversary uses no distributional information alongside the results from above, where we do use information about the bid distribution.

\begin{table}[ht!]
    \centering
    \caption{\textbf{Bid recovery rates without distributional knowledge.} We show the portion of the bids (in percentage $\pm$ standard error) that the adversary can recover within a tolerance of $\pm0.02$. }
    \label{tab:app-recovery-rates-zeros}
    \footnotesize
    \begin{tabular}{llcccccccccc}
    \toprule
                        &               & \multicolumn{5}{c}{Bidders $\times$ Items} \\
                        &               & 1 $\times$ 2     & 2 $\times$ 2     & 3 $\times$ 2     & 2 $\times$ 3      & 3 $\times$ 3     \\
    \midrule
    W/ bids             &Sample $P$     & --               & $84.98 \pm 0.50$ & $82.62 \pm 0.39$ & $91.44 \pm 0.45$  & $82.97 \pm 6.58$ \\
                        &Zero Init.     & --               & $80.23 \pm 1.16$ & $60.14 \pm 1.80$ & $73.54 \pm 2.57$  & $62.25 \pm 4.17$ \\
    \midrule
    W/O  bids           &Sample $P$     & $77.92 \pm 1.47$ & $93.08 \pm 0.64$ & $78.27 \pm 0.60$ & $25.69 \pm 1.40$  & $53.53 \pm 4.15$ \\
                        &Zero Init.     & $88.63 \pm 1.16$ & $95.39 \pm 0.35$ & $77.34 \pm 0.94$ & $26.93 \pm 1.79$  & $52.22 \pm 4.06$ \\
    \bottomrule
    \end{tabular}
\end{table}

\subsection{Compute resources}
\label{sec:app-comput}

Experiments were run on an internal cluster as jobs each utilizing a single Nvidia 2080Ti GPU.  All of the training and testing took less than 840 total GPU-hours.

\end{document}